\documentclass{article}
\newcommand{\ArxivVersion}{}

\usepackage{arxiv/PRIMEarxiv}
\usepackage{fancyhdr}

\usepackage{microtype}
\usepackage{graphicx}
\usepackage{subcaption}
\usepackage{booktabs} \usepackage[normalem]{ulem} 
\usepackage{hyperref}

\usepackage{amsmath}
\usepackage{amssymb}
\usepackage{mathtools}
\usepackage{amsthm}

\usepackage[capitalize,noabbrev]{cleveref}

\theoremstyle{plain}

\theoremstyle{definition}

\theoremstyle{remark}

\usepackage{url}
\usepackage{multirow}
\usepackage{sidecap}

\usepackage{natbib}

\usepackage{amsmath,amsfonts,amssymb}
\usepackage{hyperref}
\usepackage{url}
\usepackage{float}
\usepackage{booktabs}
\usepackage{multirow}
\usepackage{graphicx}
\usepackage{subcaption}
\usepackage[normalem]{ulem}
\usepackage{placeins}
\usepackage{algorithm}
\usepackage{algpseudocode}
\usepackage{wrapfig}
\usepackage{caption}

\newcommand{\resizeTabular}[2][]{    \begingroup
    \if\relax\detokenize{#1}\relax
        \resizebox{\linewidth}{!}{            \LARGE                                    \setlength{\tabcolsep}{4pt}            #2        }    \else
        \scalebox{#1}{            \LARGE
            \setlength{\tabcolsep}{4pt}            #2        }    \fi
    \endgroup
}

\usepackage{xcolor}
\usepackage{eso-pic}
\usepackage{zref-abspage}

\newif\ifPageLimitAuditActive
\PageLimitAuditActivefalse

\newcommand{\PageLimitAuditLimit}{0}

\AddToShipoutPictureBG{  \ifPageLimitAuditActive
    \ifnum\value{abspage}>\PageLimitAuditLimit\relax
      \AtPageLowerLeft{        \color{yellow!25}        \rule{\paperwidth}{\paperheight}      }    \fi
  \fi
}

\newcommand{\PageLimitAuditEnding}{  \ifPageLimitAuditActive
        \clearpage
    \ifnum\value{abspage}>\PageLimitAuditLimit\relax
      \PackageWarningNoLine{page-limit-audit}{        Audited content occupies \arabic{abspage} pages; limit is
        \PageLimitAuditLimit
      }    \fi
    \global\PageLimitAuditActivefalse
  \fi
}

\begin{document}

\renewcommand{\citep}[1]{\cite{#1}}
\renewcommand{\citet}[1]{Ref~\cite{#1}}

\ifdefined\ArxivVersion
\title{SRHarness: A Harness for Agentic Symbolic Regression}
\author{Zihan Yu\textsuperscript{1},
Shixuan Zhou\textsuperscript{1},
Hao Huang\textsuperscript{2},
Jingtao Ding\textsuperscript{3}\thanks{Corresponding authors.},
Yong Li\textsuperscript{1}\footnotemark[1] \\
\textsuperscript{1}Department of Electronic Engineering, BNRist, Tsinghua University, Beijing, China \\
\textsuperscript{2}College of AI, Tsinghua University, Beijing, China \\
\textsuperscript{3}Department of Earth System Science, Tsinghua University, Beijing, China \\
\texttt{yuzh23@mails.tsinghua.edu.cn, \{dingjingtao, liyong07\}@tsinghua.edu.cn}
}
\maketitle
\else
\title[mode=title]{SRHarness: A Harness for Agentic Symbolic Regression}
\author[1]{Zihan Yu}
\ead{yuzh23@mails.tsinghua.edu.cn}
\author[1]{Shixuan Zhou}
\author[2]{Hao Huang}
\author[3]{Jingtao Ding}
\cormark[1]
\ead{dingjingtao@tsinghua.edu.cn}
\author[1]{Yong Li}
\cormark[1]
\ead{liyong07@tsinghua.edu.cn}
\affiliation[1]{organization={Department of Electronic Engineering, BNRist, Tsinghua University},
  city={Beijing}, country={China}}
\affiliation[2]{organization={College of AI, Tsinghua University},
  city={Beijing}, country={China}}
\affiliation[3]{organization={Department of Earth System Science, Tsinghua University},
  city={Beijing}, country={China}}
\cortext[1]{Corresponding authors.}
\fi

\begin{abstract}
Recent agentic symbolic regression approaches increasingly rely on large language models to analyze data, select scientific operations, and refine hypotheses over long search trajectories. In such systems, performance depends not only on the underlying model and search strategy, but also on the runtime infrastructure that supports scientific search. We introduce \textbf{SRHarness}, a domain-specific harness for agentic symbolic regression built around three mechanisms: composable scientific actions that provide a common interface over raw, transformed, and candidate-derived quantities; persistent scientific state that retains evaluated hypotheses and exposes compact model-facing views; and trajectory lifecycle management that coordinates continuation, branching, restart, and termination. On LLM-SRBench, SRHarness consistently improves both numerical generalization and symbolic recovery under matched LLM backbones. With DeepSeek-v4-flash-0731, it achieves 93.69\% symbolic accuracy on LSR-Transform, compared with 62.16\% for SR-Scientist, and retains 72.97\% accuracy on an anonymized variant that removes scientific descriptions and variable semantics, versus 39.64\% for SR-Scientist. Under the same DeepSeek-v4-flash-0731 backbone, SRHarness also substantially outperforms Codex (72.97\% vs.\ 20.72\%) and reaches performance comparable to Codex with GPT-5.5, while simply providing Codex with the same scientific tools does not reproduce this advantage. These results show that effective agentic symbolic regression depends not only on models or tools, but also on structured runtime support for organizing scientific actions, accumulated hypotheses, and long-horizon search.
\end{abstract}

\ifdefined\ArxivVersion
\else
\maketitle
\fi

\hypersetup{pdfauthor={Zihan Yu, Shixuan Zhou, Hao Huang, Jingtao Ding, Yong Li},
  pdfsubject={Agentic symbolic regression},
  pdfkeywords={Symbolic regression, Scientific discovery, Large language models, Agentic systems}}

\section{Introduction}

Symbolic regression (SR) aims to discover interpretable mathematical expressions that explain observed data. Traditional SR methods design algorithms, such as genetic programming, to search the combinatorial space of mathematical expressions~\citep{koza1994genetic}, while recent LLM-based approaches leverage scientific knowledge, reasoning, and code generation to guide the discovery of mathematical expressions from data \citep{ma2024sga,grayeli2024lasr,llmsr2025,guo2025srllm}. Recent agentic SR methods take this idea further, elevating the LLM from a component that proposes or modifies equations within a predefined search procedure to a controller that autonomously selects scientific operations based on intermediate evidence \citep{srscientist2026,kepleragent2026,motsr2026}. As LLMs play a more active role in long-horizon equation discovery, system behavior increasingly depends not only on the underlying model and search strategy, but also on the surrounding execution infrastructure.

This infrastructure is commonly described as an \emph{agent harness}, which has recently received increasing attention as the layer that supports and governs agent execution over extended interactions \citep{meng2026agentharness}. For symbolic regression, such a harness must support the agent in analyzing observed data through different scientific operations while maintaining an evolving space of competing scientific hypotheses. In this process, candidate equations obtained through heterogeneous operations should share consistent representations and evaluation semantics so that they can be compared directly, while remaining available to be revisited, revised, and reused as new evidence emerges. Supporting such hypothesis-centric scientific search places domain-specific requirements on the harness, including composable scientific operations, persistent hypothesis and evidence management, and lifecycle management across long-running search trajectories. Although existing agentic SR systems already implement parts of this support, these mechanisms are typically designed around their respective search strategies rather than treated as an independent layer, so comparisons across systems may conflate differences in search strategy with differences in the supporting harness \citep{zhang2026harness}.

To address these requirements, we introduce \textbf{SRHarness}, a domain-specific harness for agentic symbolic regression that provides reusable runtime support without prescribing a particular search strategy. First, \emph{composable scientific actions} separate agent-specified scientific arguments from harness-managed execution context and support expression-based views over raw, transformed, and candidate-derived quantities, allowing heterogeneous analysis, fitting, and search procedures to operate through a common interface and evaluation semantics. Second, \emph{persistent scientific state} retains evaluated hypotheses and their supporting evidence outside the transient conversation while exposing compact model-facing views of the accumulated search state. Third, \emph{trajectory lifecycle management} coordinates continuation, branching, restart, and termination while preserving useful scientific information across long-horizon search. Together, these mechanisms make runtime support explicit, separating recurring execution, state-management, and trajectory-management concerns from the scientific search logic.

We evaluate SRHarness on LLM-SRBench and find consistent improvements over representative LLM-based and agentic SR methods under matched LLM backbones. On LSR-Synth, with DeepSeek-v4-flash-0731, SRHarness reaches 78.00\%, 78.87\%, 87.09\%, and 96.00\% OOD accuracy on physics, chemistry, biology, and materials science, respectively. On the complementary LSR-Transform track, which places greater emphasis on exact symbolic recovery, SRHarness achieves 93.69\% symbolic accuracy, compared with 62.16\% for SR-Scientist under the same backbone. We further construct LSR-Transform-Anon, which removes scientific descriptions and variable semantics while preserving the numerical observations; SRHarness still achieves 72.97\% symbolic accuracy, compared with 39.64\% for SR-Scientist. To further isolate the role of the harness, we compare against Codex under the same DeepSeek-v4-flash-0731 backbone, where SRHarness achieves 72.97\% symbolic accuracy versus 20.72\% for Codex, and reaches performance comparable to Codex with GPT-5.5 (68.47\%). Providing the same scientific tools to Codex does not reproduce this advantage. Together, these results show that effective agentic symbolic regression depends not only on the underlying model or available tools, but also on structured runtime support that organizes scientific actions, accumulated hypotheses, and long-horizon search.
\section{Related Work}

\paragraph{LLM-guided symbolic regression.}
LLM-guided SR incorporates large language models as components of algorithmic search procedures. LLM-SR uses the LLM as a mutation operator, proposing new candidates from high-scoring equations within an evolutionary loop~\citep{llmsr2025}. LaSR uses LLM-based mutation and crossover on a learned concept library~\citep{grayeli2024lasr}. SGA combines LLM-based structural search with numerical parameter optimization~\citep{ma2024sga}. SR-LLM uses retrieved knowledge to generate symbolic building blocks \citep{guo2025srllm}. Other methods improve proposal context: DrSR combines data with past generations~\citep{wang2025drsr}, ProAug supplies code-based data analysis~\citep{liu2026proaug}, and IGSR uses term-level influence scores to guide LLM proposals within Monte Carlo tree search~\citep{igsr2026}. In these methods, the LLM serves as a component that proposes or modifies equations within a predefined search procedure. Recent agentic systems like SRHarness instead treat the LLM as a controller that executes a designed search strategy by autonomously choosing scientific operations based on intermediate evidence~\citep{srscientist2026,kepleragent2026,motsr2026}.

\paragraph{Agentic symbolic regression.}
Recent systems extend the LLM's role from proposing equations to controlling analysis and search. SR-Scientist lets the agent write code to analyze observations, propose equations, and refine them through evaluation feedback~\citep{srscientist2026}. KeplerAgent follows a scientist-like reasoning process, applying physics-guided tools to first identify symmetries and other physical structure to constrain equation search~\citep{kepleragent2026}. LLM-PySR focuses on configuring the numerical solver's search space~\citep{xie2026llmpysr}. MOT-SR coordinates tool-based analysis and equation generation to balance accuracy, complexity, and generalization~\citep{motsr2026}. Deliberate Evolution (DE) adapts the choice of refinement, mutation, crossover, or regeneration to the search state~\citep{deliberateevolution2026}, while A-SR focuses on coordination among specialized agents through adaptive interaction protocols~\citep{zhao2026asr}. These works primarily develop search strategies, with supporting mechanisms typically designed around their respective strategies. In contrast, SRHarness provides reusable runtime support without prescribing a particular search strategy. Its composable scientific actions, persistent scientific state, and trajectory lifecycle management separate recurring execution, state-management, and trajectory-management concerns from the scientific search logic.

\paragraph{Agent harnesses for scientific search.}
Recent studies explicitly construct harnesses to support agent execution across different tasks. \citet{zhang2025modularharness} compose perception, memory, and reasoning modules into a reusable harness for multiple gaming environments. For scientific information processing, domain-specific designs include Beaver, which combines multimodal evidence tools and staged execution for scientific curation~\citep{zhang2026beaver}. \citet{zhang2026towards} similarly organizes an evidence-based workflow, coordinating planning, research, and writing with verification for multimodal report generation. These systems tailor execution support to environmental interaction or evidence synthesis. SRHarness supports heterogeneous analysis, fitting, and search procedures on shared scientific quantities, with candidate equations sharing consistent representations and evaluation semantics.

\section{SRHarness: A Domain-Specific Harness for Symbolic Regression}
\label{sec:method}

\subsection{Problem Setup and Overview}
\label{sec:runtime-formulation}

\begin{figure*}[t]
    \centering
    \includegraphics[width=\textwidth]{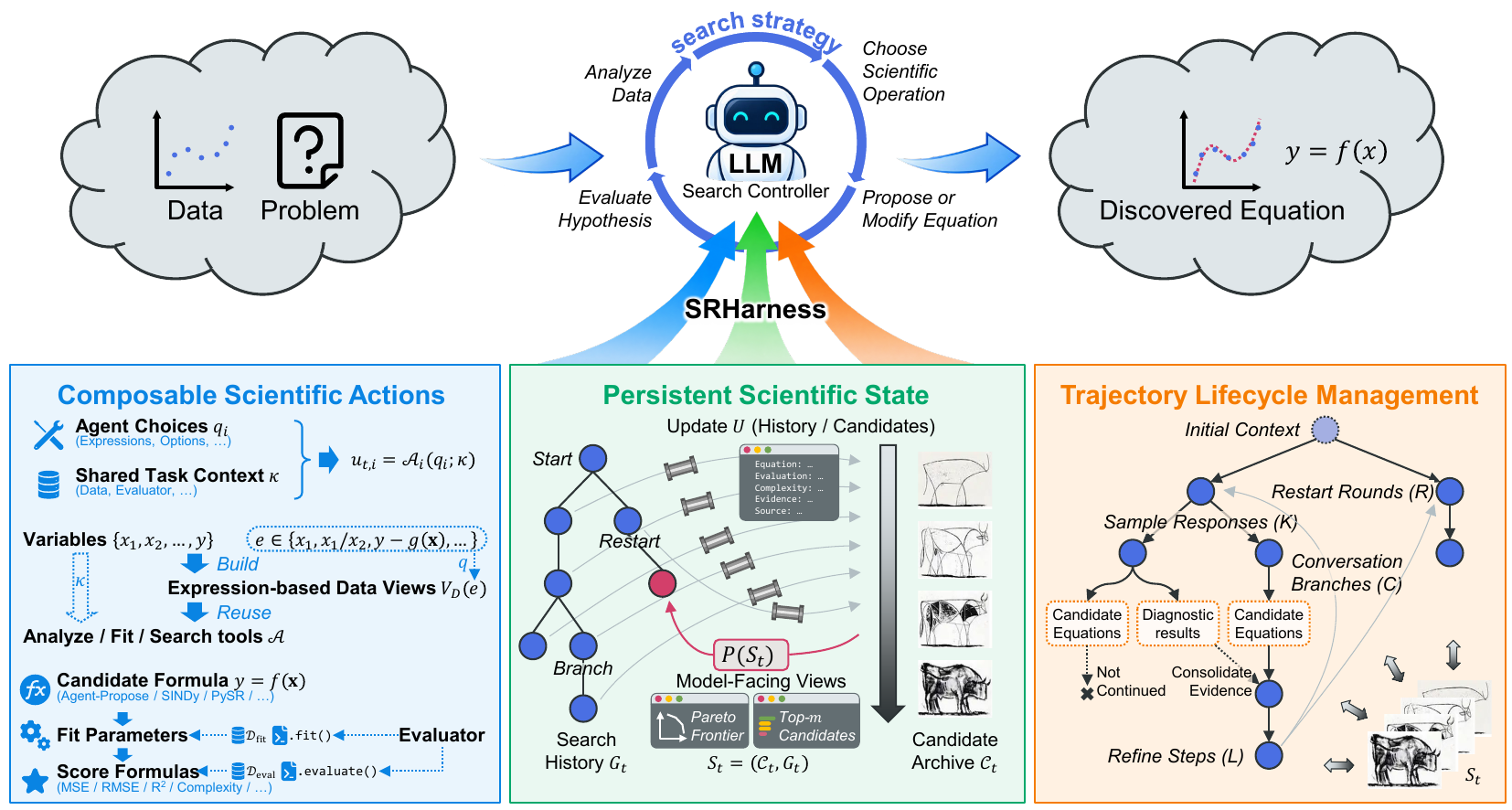}
    \caption{Overview of SRHarness.
    \textbf{Left: Composable scientific actions.} Actions combine agent choices with shared task context, reuse tools through expression-based data views, and share an evaluator for fitting and scoring candidate formulas.
    \textbf{Center: Persistent scientific state.} Search history and evaluated target equations are retained in $S_t$, with Pareto-frontier and Top-$m$ views guiding further search.
    \textbf{Right: Trajectory lifecycle management.} Response sampling, conversation branching, refinement, and restart organize discovery while consolidating evidence across trajectories.}
    \label{fig:srharness}
\end{figure*}

Given observations $D=\{(\mathbf{x}_i,y_i)\}_{i=1}^{n}$ and an optional scientific description $d$, symbolic regression seeks an explicit expression $f(\mathbf{x};\boldsymbol{\theta})$ with a suitable structure and numerical parameters. An agent chooses which analyses, transformations, and searches to perform; these choices require an execution environment that preserves their mathematical meaning and accumulated evidence. We distinguish the base model, which generates decisions, the search strategy, which organizes discovery, and the harness, which executes those decisions and maintains the search process. Following \citet{meng2026agentharness}, SRHarness supplies domain-specific execution, state management, and governance through the three mechanisms in Figure~\ref{fig:srharness}.

Let $h_t$ be the active conversation, $S_t$ the persistent scientific state, and $b_t$ the remaining search budget. The runtime context $\kappa$ binds the data, designated target, fitting and validation partitions, evaluator, and workspace. A refinement step can be written as
\begin{equation}
\begin{aligned}
 z_t &\sim \pi_M\!\left(\,\cdot\mid d,h_t,P(S_t),b_t\right),
 &
 \mathbf{u}_t &= \left\{\mathcal{A}_i(q_i;\kappa)\right\}_{i=1}^{m_t},
 \\
 S_{t+1} &= U(S_t,\mathbf{u}_t),
 &
 h_{t+1} &= \Lambda(h_t,z_t,R(\mathbf{u}_t),S_{t+1},b_t),
\end{aligned}
\label{eq:runtime-step}
\end{equation}
Here $z_t$ contains the model's text and $m_t$ tool calls. Each call $i$ selects an action $\mathcal{A}_i$ and scientific arguments $q_i$, and $\mathbf{u}_t$ collects the returned results. $U$ receives these structured results, while $R(\mathbf{u}_t)$ provides the corresponding feedback for the conversation. The operators in Eq.~\eqref{eq:runtime-step} correspond to the three harness mechanisms: \emph{composable scientific actions} $\mathcal{A}_i$ carry out the requested computations under $\kappa$; \emph{persistent scientific state} uses $U$ to store evidence and eligible candidates and $P$ to summarize archived equations and their scores for the model; and \emph{trajectory lifecycle management} uses $\Lambda$ to organize the conversation across continuation, branching, and restart. The following sections develop these mechanisms and the discovery workflow built on them.

\subsection{Composable Scientific Actions}
\label{sec:scientific-actions}

\paragraph{Scientific action interfaces.}
In symbolic regression, an agent combines relationship analysis, expression fitting, and symbolic search to turn patterns in the observations into candidate equations. Each operation requires both scientific choices and access to task data and execution settings. SRHarness separates these inputs: the model supplies expressions, candidate structures, and search options in $q_i$, while the harness supplies observations, variable bindings, and the evaluator through $\kappa$. For example, a symbolic-search call may specify an operator set in $q_i$ and obtain its data and evaluator from $\kappa$. This keeps model-generated calls focused on hypothesis exploration and allows different actions to reuse the same task setup.

Each action $\mathcal{A}_i$ takes the model-supplied arguments $q_i$ and the shared context $\kappa$, and returns
\begin{equation}
 u_{t,i}=\mathcal{A}_i(q_i;\kappa)
   =\left[s_i,\,r_i,\,\eta_i\right].
\label{eq:action-contract}
\end{equation}
Square brackets denote a result record. The execution status $s_i$ indicates success or failure, $r_i$ stores the full result, and $\eta_i$ records the tool identity and execution timing. $R_i(u_{t,i})$ formats the result or error information as feedback for the model's next decision, such as candidate equations with scores, diagnostic summaries, or an explanation of a failed call. Together, these messages form $R(\mathbf{u}_t)=\{R_i(u_{t,i})\}_{i=1}^{m_t}$ in Eq.~\eqref{eq:runtime-step}. Feedback length is capped to limit the context consumed by verbose tool outputs during long searches.

\paragraph{Expression-based data views.}
Symbolic regression often constructs a target equation by progressively simplifying dependencies in the observations through transformations and decomposition. For example, a change of coordinates can expose a simple relation, compensating for a suspected factor can reveal an invariant, and subtracting a provisional model can isolate a missing component. To support this process, SRHarness allows the same tools used to evaluate candidate formulas $y=f(\mathbf{x})$ or analyze input--target relationships to operate on expression-defined quantities.

An expression $e$, built from observed variables, constants, and supported operators with all parameters bound, specifies a virtual data column. Evaluating it on each observation in $D$ gives
\begin{equation}
 \underbrace{e(\mathbf{x},y)}_{\text{expression argument}}
 \quad\xrightarrow{\text{evaluate on }D}\quad
 \underbrace{V_D(e)=
 \{e(\mathbf{x}_i,y_i)\}_{i=1}^{n}}
 _{\text{data column supplied to the tool}}.
\label{eq:expression-view}
\end{equation}
Changing $e$ lets the statistics tool inspect raw distributions or proposed invariants, and lets relationship analysis compare original or transformed quantities. Several expression columns can jointly serve as a tool's inputs, with another providing its target. For example, after evaluating $f_0$, the agent can investigate whether its residual depends on the ratio $x_1/x_2$, then reuse these columns to fit a correction and construct a revised target equation:
\begin{equation}
\begin{gathered}
 \underbrace{\mathbf{z}=V_D(x_1/x_2)}_{\text{transformed input}},
 \qquad
 \underbrace{\mathbf{r}=V_D\!\left(y-f_0(\mathbf{x})\right)}_{\text{residual target}},
 \\[4pt]
 \mathbf{r}\approx g_{\boldsymbol{\theta}}(\mathbf{z})
 \quad\Longrightarrow\quad
 f_1(\mathbf{x})=f_0(\mathbf{x})
   +g_{\widehat{\boldsymbol{\theta}}}(x_1/x_2).
\end{gathered}
\label{eq:residual-composition}
\end{equation}
The complete $f_1$ is then evaluated against $y$. Search adapters perform the same substitution of feature expressions when returning formulas from transformed coordinates to the original variable vocabulary.

\paragraph{Unified evaluation protocol.}
To compare candidate equations $y=f(\mathbf{x})$ produced by different actions, the harness must evaluate them on the same data partitions and with the same scoring rules. For example, SINDy constructs sparse combinations of basis functions, PySR evolves expression trees, and the formula-evaluation tool can fit parameters in an expression structure proposed directly by the agent. All three use the shared evaluator to compute the scores used for candidate comparison.

The evaluator initially partitions the training observations available to the agent, $D$, into $D_{\mathrm{fit}}$ for parameter fitting and $D_{\mathrm{eval}}$ for internal evaluation. These subsets are used in all subsequent formula evaluations. Benchmark ID and OOD test sets remain reserved for evaluation after search. The protocol then proceeds in two stages: 1) When requested, the evaluator fits the expression's parameters on $D_{\mathrm{fit}}$ by minimizing mean squared error. This supports both global scalar parameters and grouped parameters $a_k$, with one coefficient shared by all observations carrying the same categorical label $k$. Tools that supply a formula with parameters already fixed can proceed directly to evaluation. 2) For every returned formula, the evaluator computes the same metric record on each subset, holding the expression and its parameters fixed. For a relation $v=f(\mathbf{x})$, where $v_i$ is the value of the target expression on observation $i$, the default record includes
\begin{equation}
\begin{aligned}
 \mathrm{MSE}_s
   &=\frac{1}{|D_s|}\sum_{(\mathbf{x}_i,y_i)\in D_s}
       \left(f(\mathbf{x}_i)-v_i\right)^2,\\
 \mathbf{m}_s
   &=\left[\mathrm{MSE}_s,\,R_s^2,\,\mathrm{RMSE}_s,\,c(f),\,\ldots\right],
   \qquad s\in\{\mathrm{fit},\mathrm{eval}\}.
\end{aligned}
\label{eq:shared-evaluation}
\end{equation}
The default complexity $c(f)$ is the number of expression-tree nodes; additional metrics include correlation measures. Scores on $D_{\mathrm{fit}}$ describe fit to the fitting observations, while scores on $D_{\mathrm{eval}}$ provide internal validation evidence for candidate selection. These common records make accuracy and complexity comparable across tools and supply the archive's admission and ranking rules (Section~\ref{sec:candidate-admission}).

For more diverse tasks, such as discovering interactions on graphs and hypergraphs or recovering dynamical laws, symbolic regression requires evaluators adapted to the data structure and scientific objective. Graph evaluation must preserve relational structure during splitting and evaluation, while dynamical models may require integration-based trajectory errors. SRHarness therefore allows task-specific data splitting, fitting, and scoring through the evaluator in $\kappa$, while retaining a common protocol across actions within each task.

\subsection{Persistent Scientific State}
\label{sec:persistent_scientific_state}
\label{sec:candidate-admission}

Long searches need to retain evaluated equations and their supporting evidence as conversations branch and restart. SRHarness maintains this information in $S_t=(\mathcal{C}_t,G_t)$: $\mathcal{C}_t$ stores admitted target candidates, and $G_t$ records model responses, tool calls, results, and their dependencies. The state update $U$ in Eq.~\eqref{eq:runtime-step} extends both components, while the projection $P$ selects evidence for subsequent model decisions.

\paragraph{Candidate admission.}
Expression-based exploration evaluates both target equations and intermediate relations. To keep archived candidates comparable, admission requires an equation $y=f(\mathbf{x})$ with fixed parameters, a right-hand side independent of $y$, and evaluation under the shared protocol. For example, fitting $\log y=g(\log(\mathbf{x}))$ on positive inputs and targets may reveal simpler dependencies, but its errors are measured on the logarithmic target scale and cannot be compared directly with those of candidates $y=f(\mathbf{x})$. The reconstructed equation $y=\exp(g(\log(\mathbf{x})))$ is therefore evaluated on the original target scale before admission.

Admission additionally requires finite MSE on each evaluated subset and finite ranking and complexity values. The candidate records in $\mathbf{u}_t$ that satisfy these requirements form the set $\mathcal{N}_t$ admitted at step $t$. Exploratory evaluations and failed calls are retained in the search history with their diagnostic results or error information to inform subsequent investigations.

\paragraph{Evidence retention and provenance.}
Each admitted record links the bound expression to its evaluation metrics, any requested residual diagnostics, and the search node that produced it. Residual summaries, large-error observations, and residual--variable correlations help identify which parts of a candidate require revision. Within the state update $S_{t+1}=U(S_t,\mathbf{u}_t)$, the archive accumulates newly admitted records as
\begin{equation}
 \mathcal{C}_{t+1}=\mathcal{C}_t\cup\mathcal{N}_t.
\label{eq:archive-update}
\end{equation}
The same update extends $G_t$ with the new execution records and their dependency links. These links record which earlier responses were continued, which candidates seeded a restart, and which tool results were merged into a later context. Together, the archive and these links preserve the evidence behind each candidate across changes to the active conversation, so previously discovered equations remain available for comparison and reuse.

\paragraph{Candidate ranking and state projection.}
To guide further search within the model's context budget, $P(S_t)$ presents selected equations and their scores from the accumulated archive. Candidates are ranked by a configured score $q(a)$, oriented so that smaller values are better, with complexity $c(a)$ breaking score ties. Our experiments use $q(a)=\mathrm{MSE}_{\mathrm{eval}}(a)$. The projection also represents the trade-off between accuracy and simplicity through the Pareto frontier. A record dominates another if it is no worse in both score and complexity and strictly better in at least one:
\begin{equation}
\begin{aligned}
 & a\prec b
 \quad \Longleftrightarrow\quad
 q(a)\leq q(b),\ c(a)\leq c(b),\ q(a)<q(b)\ \text{or}\ c(a)<c(b),\\
 & \mathcal{P}_t=\{a\in\mathcal{C}_t:\nexists b\in\mathcal{C}_t\text{ with }b\prec a\}.
\end{aligned}
\label{eq:pareto-view}
\end{equation}
During refinement, the model receives representative formulas and metrics from $\mathcal{P}_t$, exposing alternatives at different levels of complexity. At restart, the Top-$m$ ranked records provide starting hypotheses for new trajectories. These views select the evidence carried into the next conversation, while the full archive and search history remain available for later inspection and reuse.

\subsection{Trajectory Lifecycle Management}
\label{sec:trajectory-lifecycle}

Symbolic discovery requires both sustained refinement of promising hypotheses and exploration of alternative structures. The lifecycle operator $\Lambda$ in Eq.~\eqref{eq:runtime-step} organizes the active conversation across continuation, branching, and restart using tool feedback, the updated scientific state, and the remaining budget. Our reference agent combines this control with diagnostic guidance for selecting scientific actions within each trajectory.

\paragraph{Budgeted trajectory scheduling.}
The scheduler allocates $R$ restart rounds, $C$ conversation branches per round, $L$ refinement steps per branch, and $K$ response samples per step. Each round begins with a fresh prompt containing the task description and the Top-$m$ archived candidates. The $C$ branches maintain separate conversation histories while sharing $S_t$, so their findings contribute to a common pool of scientific evidence. Increasing $L$ extends refinement within a history; increasing $R$, $C$, or $K$ allocates more responses to fresh trajectories or alternative continuations. The resulting budget is at most $RCLK$ primary search-model responses. Within each branch, diagnostic evidence guides the choice of analyses, fitting operations, and hypothesis revisions.

\paragraph{Evidence-guided hypothesis refinement.}
\label{sec:discovery-workflow}
Alongside its scientific tools, SRHarness provides diagnostic guidance that suggests useful transformations, candidate families, and further analyses based on observed data patterns. Variable ranges and domain restrictions inform which transformations are appropriate, while relationship analysis tests whether derived quantities admit simpler dependencies. The relationship tool fits simple predictors on subsets of the available training observations and evaluates them on the remaining folds. Predictive scores, variation across folds, and data coverage guide the selection of a target transformation $\tau_t$, input expressions $\boldsymbol{\phi}_t$, and a function family $\mathcal{G}_t$, prioritizing simple additive, power, rational, or periodic forms before expanding the search:
\begin{equation}
 \underbrace{\tau_t(y,\mathbf{x})}_{\text{transformed target}}
 \approx
 g_{\boldsymbol{\theta}}\!\bigl(
 \underbrace{\boldsymbol{\phi}_t(\mathbf{x})}_{\text{input expressions}}
 \bigr),
 \qquad g_{\boldsymbol{\theta}}\in\mathcal{G}_t.
\label{eq:structural-hypothesis}
\end{equation}
These choices specify the quantities and expression families used by subsequent analysis, fitting, or symbolic-search actions.

Fitting turns the selected hypothesis into a target equation. For example, evidence that the compensated target $v=yx_3$ depends on the ratio $\phi=x_1/x_2$ through a power relation on positive data leads to fitting $a,p$ and reconstructing
\begin{equation}
 yx_3\approx a\left(\frac{x_1}{x_2}\right)^p
 \quad\Longrightarrow\quad
 f(\mathbf{x})=\frac{\widehat a}{x_3}
                    \left(\frac{x_1}{x_2}\right)^{\widehat p}.
\label{eq:hypothesis-reconstruction}
\end{equation}
The reconstructed equation is evaluated against $y$ under the shared protocol.

Evaluation determines the next revision: residual dependence motivates different coordinates or an additive correction (Eq.~\eqref{eq:residual-composition}); errors concentrated near a pole motivate checks of domain validity and numerical stability; and a well-fitting structure motivates tests of simpler exponents and constants. Each revision returns to evaluation of the complete formula, providing evidence to refine, simplify, or replace the current hypothesis.

\paragraph{Continuation selection and evidence sharing.}
At each refinement step, the runtime executes the tool calls from all $K$ sampled responses and applies the state update $U$ before choosing a continuation. For response $k$, $\mathcal{O}_k$ contains its returned formula evaluations with finite ranking and complexity values, including exploratory relations and target equations. The continuation is selected using the best available score in each response:
\begin{equation}
 s_k=\min_{a\in\mathcal{O}_k}q(a),\qquad
 k^\star=\operatorname*{arg\,min}_{1\leq k\leq K}s_k.
\label{eq:local-continuation}
\end{equation}
Responses without a scored formula receive $s_k=+\infty$; ties follow sample order. The selected response and its rendered tool feedback form the main continuation. Auxiliary diagnostic results from other samples can be incorporated into the same conversation, with their source links retained in the search history. Thus $\Lambda$ constructs the next conversation from the selected response, shared evidence, and remaining-budget information, after $U$ has recorded the results and admitted eligible target candidates from all samples. Algorithm~\ref{alg:rclk-scheduler} summarizes this update order.

\begin{algorithm}[t]
    \caption{$R$--$C$--$L$--$K$ scheduling with shared scientific state.}
    \label{alg:rclk-scheduler}
    \footnotesize
    \begin{algorithmic}[1]
\Require Observations $D$, task description $d$, budgets $(R,C,L,K)$, restart size $m$
\Ensure Best archived target expression and run status
\State $\kappa\gets\Call{BindContext}{D,d}$; $S=(\mathcal{C},G)\gets(\varnothing,\varnothing)$
\For{$r=1,\ldots,R$}
    \State $H\gets\Call{TopM}{\mathcal{C},m}$
    \State $h^0\gets\Call{InitialPrompt}{d,H,L}$
    \For{$c=1,\ldots,C$}
        \State $h\gets\Call{Copy}{h^0}$; $p\gets\Call{RestartLinks}{H}$
        \State $h\gets\Call{InitializeDiscovery}{h;\kappa}$ \Comment{Diagnostics and discovery skill}
        \For{$\ell=1,\ldots,L$}
            \State $z_{1:K}\gets\Call{Model}{h,P(S),\ell,L,K}$
            \For{$k=1,\ldots,K$}
                \State $\left\langle(\mathcal{A}_i,q_i)\right\rangle_{i=1}^{m_k}\gets\Call{Calls}{z_k}$
                \State $u_k\gets\left\langle\mathcal{A}_i(q_i;\kappa)\right\rangle_{i=1}^{m_k}$
            \EndFor
            \State $G\gets\Call{Record}{G,z_{1:K},u_{1:K},p,(r,c,\ell)}$
            \State $\mathcal{C}\gets\Call{AdmitAll}{\mathcal{C},u_{1:K}}$ \Comment{Eq.~\eqref{eq:archive-update}}
            \State $k^\star\gets\Call{SelectResponse}{u_{1:K}}$ \Comment{Eq.~\eqref{eq:local-continuation}}
            \State $E\gets\Call{AuxiliaryEvidence}{u_{-k^\star}}$
            \State $(h,p)\gets\Call{Continue}{h,z_{k^\star},u_{k^\star},E,P(S)}$
            \State $\sigma\gets\Call{CheckTermination}{S,(r,c,\ell);\kappa}$
            \If{$\sigma\neq\mathrm{continue}$}
                \State \Return $(\Call{Best}{\mathcal{C}},\sigma)$
            \EndIf
        \EndFor
    \EndFor
\EndFor
\State \Return $(\Call{Best}{\mathcal{C}},\mathrm{completed})$
\end{algorithmic}

\end{algorithm}

\paragraph{Termination and result preservation.}
The runtime checks termination after recording each step's results and admitting its candidates. Search ends when the response budget is exhausted or a configured stopping condition is met, such as fitting MSE falling below a specified threshold or the agent issuing an explicit stop request. Completion returns the run status, accumulated candidate records, and the best archived target expression when available. Interruption or failure preserves the recorded state in a partial result with the corresponding status, so completed evaluations remain available for inspection and reuse.

\section{Experiment}

\subsection{Experiment Setup}

\noindent\textbf{Benchmark.}
We evaluate SRHarness on LLM-SRBench \citep{llmsrbench2025}, which covers two complementary tracks with 240 problems in total. LSR-Synth contains 128 synthetically generated problems across physics, chemistry, biology, and materials science, and evaluates numerical generalization on both in-domain (ID) and out-of-domain (OOD) test sets. LSR-Transform contains 111 problems obtained by transforming established scientific equations into less familiar forms, placing greater emphasis on exact symbolic recovery. We additionally construct LSR-Transform-Anon, a variant of LSR-Transform that preserves the same numerical observations while removing scientific descriptions and anonymizing all variable names, requiring agents to discover the formula solely through data analysis rather than by utilizing scientific knowledge.

\noindent\textbf{Baselines.}
We compare against PySR \citep{pysr}, a representative conventional symbolic-regression method, and three LLM-based methods spanning different levels of agentic search: LLM-SR \citep{llmsr2025}, IGSR \citep{igsr2026}, and SR-Scientist \citep{srscientist2026}. For controlled comparisons, we rerun the LLM-based baselines using their official open-source implementations with matched backbones.

\noindent\textbf{Metrics.}
Following LLM-SRBench, we evaluate both numerical fidelity and symbolic recovery. We report pointwise accuracy $\mathrm{Acc}_{0.1}$ and normalized mean squared error (NMSE) for numerical performance, together with symbolic accuracy (SA), which measures whether the discovered expression is structurally equivalent to the reference equation. Because LSR-Transform does not provide an OOD test split, it cannot directly evaluate extrapolation, making numerical performance on this track more susceptible to overfitting. We therefore additionally report expression complexity on LSR-Transform and LSR-Transform-Anon to help distinguish compact laws from overly complex surrogate fits. We additionally report wall-clock time, model-token usage, and API cost where available.

\noindent\textbf{Evaluation Protocol.}
For each problem, the official training observations are provided to the agent, while all benchmark test observations are never exposed during search or candidate selection. Unless otherwise stated, SRHarness uses the same action set and search configuration across all experiments. All locally reproduced methods are evaluated using a common numerical and symbolic evaluation pipeline. Full benchmark descriptions, baseline configurations, metric definitions, and implementation details are provided in Appendix~\ref{sec:exp-setup}.

\subsection{Results and Analysis}

\subsubsection{Performance on LSR-Synth}

In Table~\ref{tab:lsr_synth_main} we summarize numerical and symbolic performance on LSR-Synth. Under both matched backbones, SRHarness achieves the strongest overall results across the more challenging physics, chemistry, and biology domains, on both ID and OOD splits. The improvement also extends to exact symbolic recovery: with DeepSeek-v4-flash-0731, SRHarness reaches 6.20\% SA compared with below 1\% for the matched LLM-based baselines, while with GLM-5.3-flash it reaches 10.85\%. This consistency across different base models supports the central motivation of SRHarness: the harness provides reusable runtime support for scientific search that is not tied to a particular backbone. Material-science problems appear to be substantially easier, with most methods already approaching numerical saturation; our diagnostics show that many material laws are well approximated by low-degree polynomials over the sampled ranges (Appendix~\ref{sec:lsrsynth-domain-difficulty}).

A second notable observation is the search efficiency of agentic methods. Unlike LLM-SR and IGSR, which execute largely predefined iterative search procedures, SRHarness allows the LLM to act as the controller and adaptively decide which scientific operation to perform next. With DeepSeek-v4-flash-0731, SRHarness requires 11.11 minutes per problem on average, compared with 24.02 minutes for IGSR and 57.67 minutes for LLM-SR. SR-Scientist, which similarly gives the LLM substantially more control over the search process, is also markedly faster than these predefined-search baselines. This suggests that adaptive agentic control can achieve strong discovery performance without requiring longer search time.

Varying the backbone within the same SRHarness reveals a clear but non-monotonic model dependence: larger or nominally stronger models do not consistently perform better on LSR-Synth. An exploratory comparison across the five fully evaluated backbones further shows that SR performance aligns more closely with long-horizon agent benchmarks than with model scale or scientific question-answering scores (Appendix~\ref{sec:backbone-analysis}). These observations suggest that performance in agentic symbolic regression depends not only on the base model's static capabilities, but also on how effectively it operates within a long-horizon, feedback-driven search process, highlighting the interaction between the model and the harness. Domain-level NMSE and resource usage are reported in Appendix~\ref{sec:lsr-synth-additional-results}.

\begin{table*}[h]
    \centering
    \caption{Performance on LSR-Synth across domains, reporting ID/OOD numerical accuracy, symbolic accuracy (SA), and average search time.}
    \label{tab:lsr_synth_main}
    \resizeTabular{\begin{tabular}{llcccccccccccccc}
\toprule
\multirow{2}[4]{*}{\textbf{Base Model}} & \multirow{2}[4]{*}{\textbf{Method}} & \multicolumn{2}{c}{\textbf{Physics (Acc0.1↑)}} &       & \multicolumn{2}{c}{\textbf{Chemistry (Acc0.1↑)}} &       & \multicolumn{2}{c}{\textbf{Biology (Acc0.1↑)}} &       & \multicolumn{2}{c}{\textbf{Material (Acc0.1↑)}} &       & \multicolumn{2}{c}{\textbf{Overall}} \\
\cmidrule{3-4}\cmidrule{6-7}\cmidrule{9-10}\cmidrule{12-13}\cmidrule{15-16}      &       & \textbf{ID} & \textbf{OOD} &       & \textbf{ID} & \textbf{OOD} &       & \textbf{ID} & \textbf{OOD} &       & \textbf{ID} & \textbf{OOD} &       & \textbf{SA (\%) ↑} & \textbf{Time (min)} \\
\midrule
(no LLM) & PySR  & 86.83\% & 82.87\% &       & 92.49\% & 66.68\% &       & 75.52\% & 59.26\% &       & 99.90\% & 100.00\% &       & 6.20\% & 25.46 \\
\midrule
\multirow{4}[2]{*}{Deepseek-v4-flash-0731} & LLM-SR & 71.41\% & 60.53\% &       & 72.02\% & 47.52\% &       & 68.78\% & 55.30\% &       & 95.00\% & 92.10\% &       & 0.00\% & 57.67 \\
      & IGSR  & 72.37\% & 61.92\% &       & \uline{91.71\%} & \uline{76.14\%} &       & \uline{86.95\%} & \uline{71.84\%} &       & \textbf{98.89\%} & \textbf{100.00\%} &       & \uline{0.78\%} & 24.02 \\
      & SR-Scientist & \uline{77.48\%} & \uline{69.83\%} &       & 78.08\% & 60.63\% &       & 70.17\% & 60.43\% &       & \uline{95.76\%} & \uline{97.19\%} &       & \uline{0.78\%} & \uline{16.62} \\
      & \textbf{Ours} & \textbf{84.46\%} & \textbf{78.00\%} &       & \textbf{95.24\%} & \textbf{78.87\%} &       & \textbf{95.14\%} & \textbf{87.09\%} &       & 95.06\% & 96.00\% &       & \textbf{6.20\%} & \textbf{11.11} \\
\midrule
\multirow{4}[2]{*}{GLM-5.3-flash} & LLM-SR & 67.71\% & 54.56\% &       & \uline{92.22\%} & \uline{78.06\%} &       & \uline{74.09\%} & \uline{62.30\%} &       & 96.24\% & \uline{97.45\%} &       & 0.78\% & 23.06 \\
      & IGSR  & \uline{68.33\%} & \uline{58.84\%} &       & 89.41\% & 74.18\% &       & 73.27\% & 54.82\% &       & 95.24\% & \textbf{97.48\%} &       & 1.55\% & \textbf{1.43} \\
      & SR-Scientist & 63.23\% & 56.00\% &       & 60.80\% & 48.42\% &       & 68.36\% & 59.05\% &       & \uline{96.74\%} & 96.28\% &       & \uline{6.98\%} & 16.72 \\
      & \textbf{Ours} & \textbf{88.84\%} & \textbf{85.55\%} &       & \textbf{97.08\%} & \textbf{86.17\%} &       & \textbf{88.10\%} & \textbf{79.61\%} &       & \textbf{96.86\%} & 96.00\% &       & \textbf{10.85\%} & \uline{13.07} \\
\midrule
Deepseek-v4-flash-0731 & \multirow{5}[2]{*}{\textbf{Ours}} & 84.46\% & 78.00\% &       & 95.24\% & 78.87\% &       & 95.14\% & \uline{87.09\%} &       & 95.06\% & 96.00\% &       & 6.20\% & \uline{11.11} \\
Deepseek-v4.1-flash &       & \uline{92.68\%} & \uline{89.67\%} &       & 93.94\% & \textbf{86.36\%} &       & \textbf{98.44\%} & \textbf{98.59\%} &       & \textbf{100.00\%} & \textbf{100.00\%} &       & \textbf{15.50\%} & 41.20 \\
Deepseek-v4-pro-0813 &       & \textbf{93.73\%} & \textbf{92.18\%} &       & \uline{96.86\%} & 81.39\% &       & 89.51\% & 76.93\% &       & \uline{99.82\%} & \textbf{100.00\%} &       & \uline{11.63\%} & 16.63 \\
GLM-5.3-flash &       & 88.84\% & 85.55\% &       & \textbf{97.08\%} & \uline{86.17\%} &       & 88.10\% & 79.61\% &       & 96.86\% & 96.00\% &       & 10.85\% & 13.07 \\
GLM-5.3 &       & 87.06\% & 83.16\% &       & 96.49\% & 83.30\% &       & \uline{95.99\%} & 86.51\% &       & 97.97\% & \uline{99.36\%} &       & 10.85\% & \textbf{7.48} \\
\bottomrule
\end{tabular}}
\end{table*}

\subsubsection{Performance on LSR-Transform}

We next evaluate SRHarness on LSR-Transform, where successful discovery requires recovering transformed scientific relationships rather than merely fitting the observations. As shown in Table~\ref{tab:lsr_transform}, SRHarness achieves the strongest symbolic recovery under both matched backbones, reaching 93.69\% SA with DeepSeek-v4-flash-0731 and 84.68\% with GLM-5.3-flash, while retaining strong numerical accuracy and yielding the lowest-complexity expressions among matched LLM-based methods. Notably, numerical accuracy alone poorly distinguishes symbolic discovery quality. Under DeepSeek-v4-flash-0731, all matched LLM-based methods reach NMSE on the order of $10^{-14}$, yet their SA differs substantially. LLM-SR and IGSR obtain average expression complexities of 64.46 and 42.14, respectively, compared with 14.64 for SRHarness. Thus, near-perfect numerical fitting can be achieved by substantially more complex surrogate expressions without recovering the intended symbolic structure, whereas SRHarness more consistently identifies compact expressions that are structurally equivalent to the target equations.

This distinction is important for scientific equation discovery: the objective is not merely to interpolate the observed data, but to identify a concise mathematical law underlying them. The LSR-Transform results therefore suggest that SRHarness improves symbolic recovery while maintaining strong numerical fidelity. Additional backbones show the same behavior; Appendix~\ref{sec:lsr-transform-additional-results} reports detailed efficiency and backbone comparisons.

\begin{table*}[h]
    \centering
    \caption{Performance on LSR-Transform, reporting numerical performance, symbolic accuracy (SA), expression complexity, and average search time.}
    \label{tab:lsr_transform}
    \resizeTabular[0.37950008]{\begin{tabular}{llccccc}
\toprule
\multirow{2}[4]{*}{\textbf{Base Model}} & \multirow{2}[4]{*}{\textbf{Method}} & \multicolumn{5}{c}{\textbf{LSR-Transform}} \\
\cmidrule{3-7}      &       & \textbf{Test Acc0.1 (\%)↑} & \textbf{Test NMSE↓} & \textbf{SA (\%) ↑} & \textbf{Complexity↓} & \textbf{Time (min)} \\
\midrule
(no LLM) & PySR  & 77.33\% & 1.44E-04 & 35.14\% & 17.57 & 25.49 \\
\midrule
\multirow{4}[0]{*}{Deepseek-v4-flash-0731} & LLM-SR & 79.48\% & 2.80E-14 & 45.05\% & 64.46 & 72.74 \\
      & IGSR  & \uline{86.44\%} & \textbf{1.05E-14} & 48.65\% & 42.14 & 23.67 \\
      & SR-Scientist & 79.62\% & 2.50E-14 & \uline{62.16\%} & \uline{17.50} & \textbf{6.72} \\
      & \textbf{Ours} & \textbf{88.37\%} & \uline{1.24E-14} & \textbf{93.69\%} & \textbf{14.64} & \uline{12.47} \\
\midrule
\multirow{4}[0]{*}{GLM-5.3-flash} & LLM-SR & 78.45\% & 2.50E-14 & 45.95\% & 45.19 & 27.56 \\
      & IGSR  & 52.00\% & 3.77E-02 & 20.72\% & 42.30 & \textbf{1.58} \\
      & SR-Scientist & \uline{86.38\%} & \uline{2.41E-14} & \uline{66.67\%} & \uline{17.62} & 7.41 \\
      & \textbf{Ours} & \textbf{86.99\%} & \textbf{1.50E-14} & \textbf{84.68\%} & \textbf{15.83} & \uline{7.09} \\
\midrule
Deepseek-v4-flash-0731 & \multirow{4}[0]{*}{\textbf{Ours}} & \uline{88.37\%} & \uline{1.24E-14} & \textbf{93.69\%} & \uline{14.64} & 12.47 \\
Deepseek-v4-pro &       & 86.55\% & 1.27E-14 & 84.68\% & 19.01 & 13.77 \\
GLM-5.3-flash &       & 86.99\% & 1.50E-14 & 84.68\% & 15.83 & \uline{7.09} \\
GLM-5.3 &       & \textbf{89.91\%} & \textbf{1.24E-14} & \uline{90.09\%} & \textbf{14.34} & \textbf{3.57} \\
\bottomrule
\end{tabular}}
\end{table*}

\subsection{Equation Discovery without Semantic Priors}

Although LSR-Transform was designed to reduce direct memorization by rewriting familiar scientific equations into less common forms, it still preserves the original scientific context and variable semantics. Prior work suggests that these cues can provide strong signals for recovering transformed laws, even without access to numerical observations \citep{srscientist2026, ICLR2026_a0e1c2c4}. To address this issue, we construct LSR-Transform-Anon as a paired evaluation setting: the numerical observations and target equations are unchanged, while problem descriptions and variable semantics are removed (see Appendix~\ref{sec:anonymization-details} for the exact anonymization procedure). This setting more directly probes whether an agent can use scientific tools to interrogate numerical observations, identify mathematical patterns, and iteratively recover the underlying equation.

Table~\ref{tab:lsr_transform_anon} shows that anonymization substantially degrades all LLM-based methods, confirming that semantic priors contribute to equation recovery. The extent of this degradation, however, differs sharply across search paradigms. LLM-SR and IGSR each retain only 28\% of their original symbolic accuracy, whereas SR-Scientist retains 64\% and SRHarness retains 78\%. After semantic cues are removed, LLM-SR and IGSR even fall below the non-LLM PySR baseline in symbolic accuracy, while both agentic methods remain substantially ahead of it. SRHarness is strongest in this setting, reaching 72.97\% SA while maintaining near-machine-precision numerical fit. Notably, its search time remains essentially unchanged after anonymization, whereas several baselines require substantially longer trajectories while recovering far fewer target equations (see Appendix~\ref{sec:lsr-transform-anon-additional-results} for the full resource comparison). Together, these results suggest that SRHarness's advantage is not solely attributable to exploiting pretrained scientific knowledge. Even when problem-specific semantic cues are removed, its structured support for data analysis, hypothesis testing, and iterative search enables substantially stronger recovery of mathematical structure from numerical observations.

\begin{table*}[h]
    \centering
    \caption{Performance on LSR-Transform-Anon after removing problem-specific semantic cues, reporting numerical performance, symbolic accuracy (SA), SA retention, expression complexity, and search efficiency.}
    \label{tab:lsr_transform_anon}
    \resizeTabular[0.37950008]{\begin{tabular}{llccccccc}
\toprule
\multirow{2}[4]{*}{\textbf{Base Model}} & \multirow{2}[4]{*}{\textbf{Method}} & \multicolumn{7}{c}{\textbf{LSR-Transform (Anonymized)}} \\
\cmidrule{3-9}      &       & \textbf{Test Acc0.1 (\%)↑} & \textbf{Test NMSE↓} & \textbf{SA (\%) ↑} & \textbf{SA Retention (\%)↑} & \textbf{Complexity↓} & \textbf{Time (min)↓} & \textbf{Time Ratio (×)↓} \\
\midrule
(no LLM) & \multicolumn{1}{l}{PySR} & 77.33\% & 1.44E-04 & 35.14\% & 100\% & 17.57 & 25.49 & 1 \\
\midrule
\multirow{4}[2]{*}{Deepseek-v4-flash-0731} & \multicolumn{1}{l}{LLM-SR} & 67.59\% & 1.21E-03 & 12.61\% & 28\%  & 87.85 & 286.13 & 3.93 \\
      & \multicolumn{1}{l}{IGSR} & 47.36\% & 8.36E-02 & 13.51\% & 28\%  & 34.05 & 55.60 & \uline{2.35} \\
      & \multicolumn{1}{l}{SR-Scientist} & \uline{73.74\%} & \uline{6.00E-05} & \uline{39.64\%} & \uline{64\%} & \textbf{30.73} & \uline{22.30} & 3.32 \\
      & \multicolumn{1}{l}{\textbf{Ours}} & \textbf{91.69\%} & \textbf{1.24E-14} & \textbf{72.97\%} & \textbf{78\%} & \uline{33.36} & \textbf{12.34} & \textbf{0.99} \\
\midrule
Deepseek-v4-flash-0731 & \multirow{3}[2]{*}{\textbf{Ours}} & \textbf{91.69\%} & \textbf{1.24E-14} & \textbf{72.97\%} & \textbf{78\%} & \textbf{33.36} & \textbf{12.34} & \textbf{0.99} \\
Deepseek-v4-pro &       & \uline{86.64\%} & \textbf{1.24E-14} & \uline{65.77\%} & \uline{78\%} & \uline{41.06} & 17.30 & \uline{1.26} \\
GLM-5.3-flash &       & 86.33\% & \uline{2.47E-14} & 59.46\% & 70\%  & 41.65 & \uline{15.15} & 2.14 \\
\bottomrule
\end{tabular}}
\end{table*}

\subsection{Harness Design Beyond Models and Tools}

To disentangle the contribution of the harness from that of the base model, we compare SRHarness with Codex under the same DeepSeek-v4-flash-0731 backbone. As reported in Table~\ref{tab:codex}, SRHarness achieves 72.97\% symbolic accuracy, compared with 20.72\% for Codex. Since both systems use the same underlying model, this gap highlights the substantial role of the surrounding runtime system in agentic symbolic regression. Notably, SRHarness with DeepSeek-v4-flash-0731 reaches comparable symbolic recovery to Codex with GPT-5.5, which achieves 68.47\% SA on the same anonymized benchmark. These results suggest that harness design and base-model capability are complementary sources of performance: structured domain-specific support can substantially amplify the capability of the underlying model. Appendix~\ref{sec:codex-anon-additional-results} reports the corresponding resource usage.

Moreover, simply providing the same scientific tools is not sufficient to reproduce the benefits of the harness. Exposing the scientific tools of SRHarness directly to Codex with GPT-5.5 decreases symbolic accuracy from 68.47\% to 64.86\%, while increasing runtime and token consumption. This suggests that the value of scientific actions depends not only on their availability, but also on how they are integrated with persistent state, evaluation, and trajectory management throughout the search. More broadly, these comparisons show that an effective SR harness is not merely a collection of tools around an LLM, but a runtime system that organizes model decisions, scientific actions, accumulated hypotheses, and feedback into structured scientific search.

\begin{table*}[h]
    \centering
    \caption{Comparison of SRHarness and Codex on LSR-Transform-Anon under different model and tool configurations.}
    \label{tab:codex}
    \resizeTabular[0.37950008]{\begin{tabular}{rlcccc}
\toprule
\multicolumn{1}{r}{\multirow{2}[4]{*}{\textbf{Base Model}}} & \multirow{2}[4]{*}{\textbf{Harness}} & \multicolumn{4}{c}{\textbf{LSR-Transform (Anonymized)}} \\
\cmidrule{3-6}      &       & \textbf{Test Acc0.1 (\%)↑} & \textbf{Test NMSE↓} & \textbf{SA (\%) ↑} & \textbf{Complexity↓} \\
\midrule
\multicolumn{1}{r}{\multirow{2}[1]{*}{Deepseek-v4-flash-0731}} & \textbf{Ours} & \textbf{91.69\%} & \textbf{1.24E-14} & \textbf{72.97\%} & \textbf{33.36} \\
      & Codex & 66.74\% & 1.59E-02 & 20.72\% & 91.72 \\
\midrule
\multicolumn{1}{r}{\multirow{2}[1]{*}{GPT-5.5}} & Codex & \textbf{93.20\%} & \textbf{1.63E-14} & \textbf{68.47\%} & 2547.55 \\
      & Codex (+tools) & 91.45\% & 2.49E-14 & 64.86\% & \textbf{902.47} \\
\bottomrule
\end{tabular}}
\end{table*}

\subsection{Ablation Study}
\label{sec:ablation}

We analyze three design dimensions of SRHarness on LSR-Transform-Anon: the scientific action space, the model-facing view of persistent scientific state, and trajectory lifecycle management. We use DeepSeek-v4-flash-0731 throughout and take the three-run average of the full configuration as the reference (see Appendix Table~\ref{tab:ablation-repeat} for the individual runs). The ablation results are summarized in Figure~\ref{fig:ablation}, which plots symbolic accuracy against mean cumulative token usage for each variant (see Appendix~\ref{sec:ablation-details} for complete variant definitions and additional metrics).

\begin{figure*}[t]
    \centering
    \includegraphics[width=\textwidth]{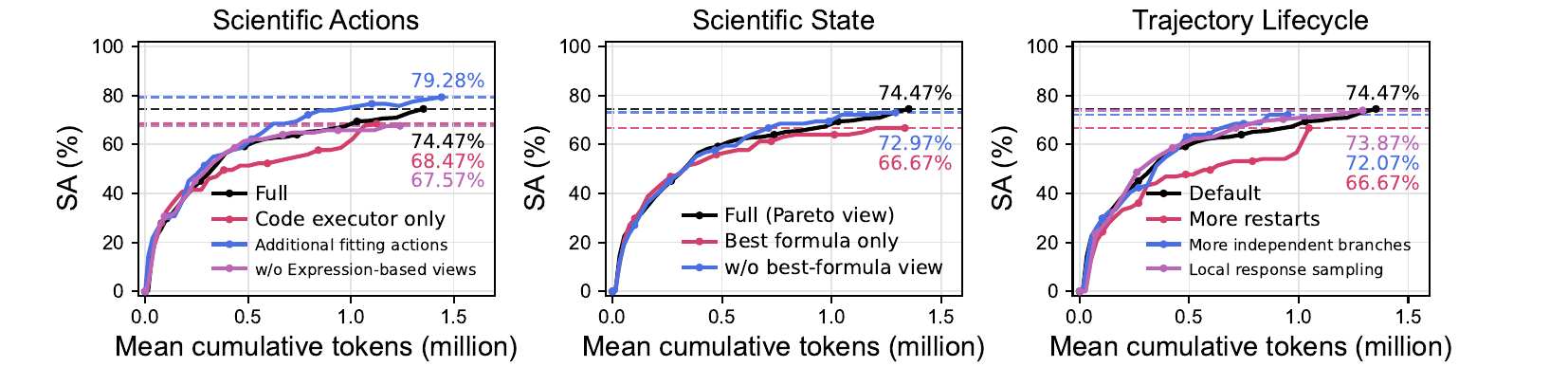}
    \caption{Ablation results on LSR-Transform-Anon. Curves show symbolic accuracy (SA) against mean cumulative token usage; dashed lines indicate final SA.}
    \label{fig:ablation}
\end{figure*}

\paragraph{Scientific action space.}
We first examine the role of the scientific action interface. Restricting SRHarness to the code executor only reduces symbolic accuracy from 74.47\% to 68.47\%, showing that structured scientific actions provide useful support beyond unrestricted code execution. We further disable expression-based views by restricting action inputs to raw dataset variables rather than transformed or candidate-derived quantities; this reduces SA to 67.57\%, indicating that composability over intermediate scientific quantities is an important part of the action interface. Conversely, adding further fitting actions improves SA to 79.28\%, demonstrating that new scientific capabilities can be incorporated through the same standardized interface and contribute to symbolic discovery (see Appendix Table~\ref{tab:ablation-action}).

\paragraph{Model-facing view of scientific state.}
We next study how accumulated hypotheses should be presented back to the model. Compared with the 74.47\% SA achieved by the Pareto-based view, presenting only the single best formula reduces symbolic accuracy to 66.67\%. Interestingly, removing the explicit best-formula view still achieves 72.97\%, only slightly below the full configuration. One possible explanation is that the Pareto view already exposes strong candidates while preserving structurally diverse alternatives, whereas repeatedly emphasizing a single incumbent may bias subsequent refinement toward an early solution. The much larger degradation from replacing the Pareto view with a single-best view than from removing the explicit best-formula view suggests that the Pareto representation provides the primary benefit, while the best-formula signal plays a secondary role (see Appendix Table~\ref{tab:ablation-state}).

\paragraph{Trajectory lifecycle management.}
Finally, we study how the search budget is allocated across restart, independent branching, refinement depth, and local response sampling. Relative to the default $R1$-$C1$-$L30$-$K1$ configuration at 74.47\% SA, reallocating refinement steps to additional restarts ($L\!\rightarrow\!15$, $R\!\rightarrow\!2$) reduces SA to 66.67\%, while additional independent branches reach 72.07\% and local response sampling reaches 73.87\%. These results show that trajectory organization itself affects discovery performance: among the tested alternatives, preserving longer refinement trajectories is more effective than allocating the same nominal budget to additional restarts or branches, while local response sampling remains closest to the default configuration. This suggests that, under the current search budget, maintaining sufficient refinement depth should take priority over increasing trajectory-level diversity, with local diversification imposing a smaller penalty than restarting or spawning additional long-horizon branches (see Appendix Table~\ref{tab:ablation-lifecycle}).

\section{Conclusion}
\label{sec:conclusion}

We introduced SRHarness, a domain-specific harness for agentic symbolic regression that separates reusable runtime support from the underlying model and search strategy. Through composable scientific actions, persistent scientific state, and trajectory lifecycle management, SRHarness consistently improves numerical and symbolic equation discovery across matched backbones and remains effective when problem-specific semantic cues are removed. Comparisons with Codex further show that these gains cannot be explained by the base model or tool availability alone: under the same backbone, SRHarness achieves substantially stronger symbolic recovery, while providing Codex with the same scientific tools does not reproduce the advantage. Together, these results establish the harness as a first-class component of agentic symbolic regression, organizing model decisions, scientific actions, accumulated hypotheses, and long-horizon search into an effective discovery process.

\FloatBarrier
\section*{CRediT authorship contribution statement}
\textbf{Zihan Yu:} Conceptualization, Investigation, Methodology, Project administration, Validation, Visualization, Writing -- original draft, Writing -- review \& editing.
\textbf{Shixuan Zhou:} Investigation, Validation, Writing -- original draft, Writing -- review \& editing.
\textbf{Hao Huang:} Methodology, Writing -- original draft, Writing -- review \& editing.
\textbf{Jingtao Ding:} Conceptualization, Project administration, Supervision, Writing -- review \& editing.
\textbf{Yong Li:} Conceptualization, Project administration, Supervision, Writing -- review \& editing.

\section*{Declaration of competing interest}
The authors declare no competing interests.

\section*{Code availability}
The code is available at \url{https://github.com/tsinghua-fib-lab/SRHarness}. Run configurations are detailed in Appendix~\ref{sec:implementation-details}.

\section*{Declaration of generative AI and AI-assisted technologies in the manuscript preparation process}

During the preparation of this work, the author(s) used ChatGPT for grammar checking and manuscript polishing. The author(s) reviewed and edited the output as needed and take full responsibility for the content of the published article.

\bibliography{reference}

\clearpage
\appendix
\ifdefined\ArxivVersion\else
\AOAppendixtrue
\fi
\counterwithin{equation}{section}
\counterwithin{figure}{section}
\counterwithin{table}{section}

\section{Experiment Setup}
\label{sec:exp-setup}

\subsection{Benchmark}

We evaluate SRHarness on LLM-SRBench~\citep{llmsrbench2025}, a benchmark designed to evaluate scientific equation discovery with LLM-based methods. Our evaluation contains 240 problems in two complementary tracks: 128 LSR-Synth problems and 111 LSR-Transform problems. We additionally evaluate an anonymized view of the same 111 LSR-Transform problems to separate discovery from numerical observations from the use of semantic scientific priors. The anonymized track is a different agent-facing condition, not an additional set of numerical problems.

\noindent\textbf{LSR-Synth.}
LSR-Synth contains 128 synthetically generated equation-discovery problems spanning physics (44), chemistry (36), biology (24), and materials science (25). Unlike benchmarks based primarily on well-known equations, LSR-Synth emphasizes discovery from numerical observations. Each problem has an official training set, an in-domain (ID) test set, and an out-of-domain (OOD) test set. We report ID and OOD numerical performance separately for each domain, while symbolic accuracy and resource usage are aggregated over all 128 problems.

\noindent\textbf{LSR-Transform.}
LSR-Transform contains 111 problems derived by transforming established scientific equations into less familiar mathematical representations. The transformations preserve the underlying relationship while reducing the usefulness of directly recalling a canonical equation from model pretraining. Compared with LSR-Synth, these problems retain meaningful scientific descriptions and variable semantics, and therefore jointly test data-driven reasoning and the ability of LLMs to exploit relevant scientific priors.

\noindent\textbf{LSR-Transform-Anon.}
\label{sec:anonymization-details}
To further isolate active structure discovery from semantic recall, we construct an anonymized agent-facing view of LSR-Transform. The target is renamed $y$, inputs are renamed $x_1,x_2,\ldots$, and scientific variable descriptions are replaced by generic labels such as ``target variable'' and ``input variable 1.'' The original variable names, descriptions, and reference expression are not included in the agent prompt or tool context; only the generic input/output roles and the unchanged numerical observations remain. The reference expression is renamed consistently and retained only by the evaluator. Standard and anonymized LSR-Transform therefore use identical training and test values, and differ only in the semantic information available to the method.

\subsection{Baselines}

We compare SRHarness against representative methods spanning conventional symbolic regression, LLM-guided search, and autonomous symbolic-regression agents. For controlled comparisons, we rerun methods with matched LLM backbones whenever their implementations permit. PySR provides a non-LLM reference, while LLM-SR, IGSR, and SR-Scientist are compared primarily under DeepSeek-v4-flash-0731 and GLM-5.3-flash. Rows marked with an asterisk are the corresponding paper's own reported results under a different backbone or budget; they are included for context but are not treated as controlled comparisons. We do not copy a paper's measurements of other methods into these rows.

\noindent\textbf{PySR.}\citep{pysr}
We include PySR as a representative non-LLM symbolic-regression baseline. PySR performs equation search using an evolutionary symbolic-regression procedure and, therefore, provides a reference point for comparing agentic approaches against a mature conventional SR system.

\noindent\textbf{LLM-SR}~\citep{llmsr2025}
is an LLM-based symbolic-regression method that iteratively proposes and evolves candidate programs within a predefined optimization workflow. It represents the class of methods in which the LLM serves primarily as a candidate generator or modifier, while the surrounding search procedure determines how candidates are evaluated and selected.

We reproduce LLM-SR using its official island-model evolutionary search framework, where the LLM proposes candidate program skeletons and learnable constants are fitted to training observations via BFGS (scipy.optimize.minimize). Under our controlled comparison, we use a budget of 200 LLM samples per problem rather than the 1000 samples used in the original LLM-SRBench protocol. Because LLM-SR's prompts are much shorter than our agent's full-dialogue context, a call-count-aligned budget would over-budget LLM-SR in tokens; we instead align on a token basis, so the reduced budget does not disadvantage the baseline. Backbones are matched to \texttt{deepseek/deepseek-v4-flash} and \texttt{z-ai/glm-5.3-flash} via OpenRouter. Both use \texttt{temperature=1.0}; chain-of-thought reasoning is disabled for DeepSeek (\texttt{reasoning: \{enabled: false\}}) and set to low effort for GLM-5.3-flash (\texttt{reasoning: \{effort: low\}}). Evaluation follows the same unified protocol as all baselines in this work.\footnote{The LLM-SRBench paper reports \texttt{temperature=0.8} and \texttt{num\_evaluators=4}, while the official implementation hard-codes \texttt{temperature=1.0} and defaults to \texttt{num\_evaluators=1}. We follow the implementation values to ensure reproducibility.}

\noindent\textbf{IGSR}~\citep{igsr2026}
incorporates LLM reasoning into the iterative symbolic search and represents a recent strong LLM-based equation-discovery approach. We include IGSR to compare SRHarness with methods that make more extensive use of LLM-guided search while retaining a largely predefined optimization procedure.

\noindent\textbf{SR-Scientist}~\citep{srscientist2026}
is a representative agentic symbolic-regression method that elevates the LLM from an equation proposer to an autonomous search controller. The agent can analyze data through code execution, evaluate candidate equations, and refine its hypotheses over long interaction trajectories. It therefore provides the most direct baseline for evaluating whether the domain-specific runtime support provided by SRHarness improves autonomous symbolic discovery.

We adopt the official open-source implementation of SR-Scientist~\citep{srscientist2026} with its default toolset and agentic loop unchanged. For controlled comparison, backbones are matched to \texttt{deepseek-v4-flash} and \texttt{glm-5.3-flash} via OpenRouter (reasoning disabled). We cap each problem at 75 LLM calls (3 outer turns $\times$ 25 inner rounds), which deviates from the 1000-call budget in the original paper. Because SR-Scientist's agentic loop accumulates full dialogue history and tool outputs across calls, a call-count-aligned budget would over-budget it in tokens; we instead align on a token basis, holding SR-Scientist within the same token order of magnitude as our method, so the reduced budget does not disadvantage the baseline. We also evaluate on both LSR-Synth and LSR-Transform rather than LSR-Synth alone. Consequently, these numbers are not directly comparable to the original paper, but they use the unified symbolic-accuracy criterion (\texttt{get\_symbolic\_acc}) shared across all baselines in this work.

All locally reproduced methods are evaluated from their archived submitted expression or algorithm-native prediction artifact using the same test data and numerical evaluator. Method-specific adapters are used when a submission is an executable program rather than a formula; when the original program did not persist its fitted parameters, we refit those parameters on the official training data before extracting and evaluating the formula. This avoids silently assigning different numerical semantics to different methods while preserving each method's native prediction semantics as closely as its artifacts permit.

\subsection{Metrics}

Following LLM-SRBench, we evaluate discovered equations from two complementary perspectives: \emph{numerical fidelity} and \emph{symbolic recovery}. Numerical metrics measure whether the discovered expression reproduces the observed relationship, whereas symbolic accuracy evaluates whether it recovers the underlying mathematical law.

\noindent\textbf{Pointwise accuracy ($\mathrm{Acc}_{0.1}$).}
For a problem with $n$ valid test predictions, we compute the fraction of test points whose absolute error is at most 10\% of the magnitude of the target:
\begin{equation}
\mathrm{Acc}_{0.1} = \frac{1}{n}\sum_{i=1}^{n}
\mathbb{I}\!\left[|\hat y_i-y_i|\leq 0.1|y_i|\right].
\end{equation}
We average this per-problem fraction across problems and report it as a percentage. Following the benchmark implementation, test points with a NaN prediction are excluded and the retained point count is recorded for each problem. Thus, our $\mathrm{Acc}_{0.1}$ is not the stricter $\mathrm{AccAll}_{0.1}$ criterion that declares a problem successful only when every test point passes the threshold. For LSR-Synth, it is evaluated separately on the ID and OOD test sets.

\noindent\textbf{Normalized Mean Squared Error (NMSE).}
For each problem, we additionally compute
\begin{equation}
\mathrm{NMSE} = \frac{\frac{1}{n}\sum_i(\hat y_i-y_i)^2}
{\frac{1}{n}\sum_i(y_i-\bar y)^2},
\end{equation}
which captures the magnitude of numerical prediction errors beyond the binary threshold used by $\mathrm{Acc}_{0.1}$. Lower values indicate better numerical agreement. We report the median NMSE across problems to reduce sensitivity to a small number of severely failed predictions.

\noindent\textbf{Symbolic Accuracy (SA).}
Symbolic accuracy measures whether the submitted expression is mathematically equivalent to the reference equation, rather than merely producing similar predictions on sampled points. We uniformly re-evaluate locally reproduced results with our symbolic-accuracy implementation. It first performs strict numerical diagnostics ($\mathrm{atol}=10^{-8}$ and $\mathrm{rtol}=10^{-6}$, including a conservative simplification of fitted constants), and then uses DeepSeek-v4-flash-0731 as a structural equivalence judge. The judge is given both expressions and the observed variable ranges and is instructed to accept algebraic rearrangements, valid domain identities, and small fitted-coefficient rounding, but to reject polynomial, rational, or Taylor surrogates for a different function family. The structural verdict determines SA; numerical closeness alone is not treated as proof of symbolic recovery. We report the mean of the resulting per-problem Boolean verdicts.

Unless noted otherwise, $\mathrm{Acc}_{0.1}$ is the mean and NMSE is the median over the available successfully archived submissions. SA instead uses the complete benchmark as its denominator (128 problems for LSR-Synth and 111 for either LSR-Transform condition), so a missing or unevaluable submission counts as symbolically incorrect. Parenthesized values marked $\dagger$ are paper-reported $\mathrm{AccAll}_{0.1}$ or mean NMSE values and therefore are not directly comparable down the corresponding column. Asterisked rows retain the reporting convention of the source paper. Time is mean wall-clock search time in minutes per problem, Token is mean model-token usage in millions, and Cost is mean recorded API cost in U.S. dollars. Unavailable or non-auditable usage are denoted by a slash (``/'').

\subsection{Implementation Details}
\label{sec:implementation-details}

For SRHarness, we use the search configuration $R1$-$C1$-$L30$-$K1$: one restart round, one global conversation branch, at most 30 refinement steps, and one local response sample at each step. The experiment seed is 26091320. The primary DeepSeek-v4-flash-0731 runs allow 4,096 output tokens per model call, whereas the primary GLM-5.3-flash runs allow 8,192; other backbone-specific limits are recorded with their runs.

For each problem, we randomly split the official training observations into 80\% fitting data and 20\% validation data with split seed 42. Scientific actions fit candidates on the fitting portion and return both fitting and validation evidence; candidates in the persistent state are ranked by validation MSE. Neither the official ID test set nor the LSR-Synth OOD test set is exposed during search or used for candidate selection. They are accessed only after a final expression has been submitted.

Across the main SRHarness experiments, the action set is \texttt{statistics\_analysis}, \texttt{relationship\_analysis}, \texttt{read\_skill}, \texttt{evaluate\_formula}, \texttt{submit\_formula}, \texttt{constant\_fit}, \texttt{call\_pysr}, \texttt{call\_sindy}, and \texttt{code\_executor}. Before the first model response, the harness runs the same statistical and relationship diagnostics and loads the same symbolic-law-discovery skill. The action set, random split, and $R1$-$C1$-$L30$-$K1$ schedule are held fixed across the three benchmark conditions and primary backbones; model-specific response limits are the exception noted above.

For controlled method comparisons, we primarily use DeepSeek-v4-flash-0731 and GLM-5.3-flash as matched backbones. We additionally run SRHarness with larger or alternative models to measure portability across model families. Every canonical result retains a pointer and hash to its immutable raw log, while numerical metrics and SA are recomputed in a common evaluation layer. This separation allows evaluator corrections without modifying the original submission, prompt, tool trace, or usage record.

\section{Detailed Experimental Results}

\subsection{Detailed LSR-Synth Results}
\label{sec:lsr-synth-additional-results}
\label{app:exp_results}

Tables~\ref{tab:lsr-synth-nmse} and~\ref{tab:lsr-synth-budget} supplement the accuracy summary with domain-level NMSE and resource usage. Detailed numerical errors in Table~\ref{tab:lsr-synth-nmse} corroborate the main-text observation that SRHarness achieves superior numerical fidelity across the more challenging scientific domains. For instance, under DeepSeek-v4-flash-0731, SRHarness attains a physics ID NMSE of $1.48\times10^{-5}$, outperforming SR-Scientist ($1.85\times10^{-5}$) and IGSR ($1.34\times10^{-4}$), while remaining highly competitive on chemistry and biology ID splits. The material domain, as discussed in Section 4.2, exhibits near-saturation for all methods, which is consistent with the low-order functional forms of material laws over the sampled ranges.

Table~\ref{tab:lsr-synth-budget} further reveals the search efficiency and cost profile of SRHarness. It requires an average of 11.11 minutes per problem under DeepSeek-v4-flash-0731, substantially faster than LLM-SR (57.67 minutes) and IGSR (24.02 minutes). Notably, this runtime advantage is achieved with a competitive API cost (\$0.03) despite a higher total token count, owing to a favorable input-dominant token profile (see Appendix B.2 for details). These resource statistics align with the main-text finding that adaptive agentic control avoids unnecessary predefined search operations.

\begin{table*}[t]
    \centering
    \caption{Domain-level ID and OOD NMSE on LSR-Synth.}
    \label{tab:lsr-synth-nmse}
    \resizeTabular[0.37950008]{\begin{tabular}{llccccccccccc}
\toprule
\multirow{2}[4]{*}{\textbf{Base Model}} & \multirow{2}[4]{*}{\textbf{Method}} & \multicolumn{2}{c}{\textbf{Physics (NMSE↓)}} &       & \multicolumn{2}{c}{\textbf{Chemistry (NMSE↓)}} &       & \multicolumn{2}{c}{\textbf{Biology (NMSE↓)}} &       & \multicolumn{2}{c}{\textbf{Material (NMSE↓)}} \\
\cmidrule{3-4}\cmidrule{6-7}\cmidrule{9-10}\cmidrule{12-13}      &       & \textbf{ID} & \textbf{OOD} &       & \textbf{ID} & \textbf{OOD} &       & \textbf{ID} & \textbf{OOD} &       & \textbf{ID} & \textbf{OOD} \\
\midrule
(no LLM) & PySR  & 8.01E-06 & 7.57E-05 &       & 5.07E-06 & 4.59E-01 &       & 2.50E-06 & 1.17E-02 &       & 1.31E-09 & 2.56E-07 \\
\midrule
\multirow{4}[2]{*}{Deepseek-v4-flash-0731} & LLM-SR & 8.71E-05 & 1.99E-03 &       & 1.08E-04 & 9.51E-01 &       & 7.79E-06 & 3.90E-02 &       & 8.31E-08 & 4.41E-06 \\
      & IGSR  & 1.34E-04 & 1.43E-03 &       & \uline{9.28E-08} & \uline{2.84E-03} &       & \textbf{1.87E-08} & \textbf{6.84E-06} &       & \uline{1.94E-11} & \uline{1.54E-08} \\
      & SR-Scientist & \uline{1.85E-05} & \uline{3.83E-04} &       & 3.05E-05 & 8.26E-02 &       & 2.81E-05 & 1.02E-02 &       & 2.39E-09 & 5.13E-07 \\
      & \textbf{Ours} & \textbf{1.48E-05} & \textbf{8.21E-05} &       & \textbf{1.80E-08} & \textbf{8.70E-04} &       & \uline{1.75E-07} & \uline{2.25E-05} &       & \textbf{3.21E-14} & \textbf{1.91E-10} \\
\midrule
\multirow{4}[2]{*}{GLM-5.3-flash} & LLM-SR & 2.62E-04 & 8.60E-03 &       & 3.08E-06 & 2.29E-02 &       & 6.41E-06 & 6.71E-02 &       & 2.03E-08 & 4.09E-06 \\
      & IGSR  & \uline{2.02E-04} & \uline{3.04E-03} &       & \uline{4.65E-07} & \textbf{2.17E-04} &       & \uline{2.17E-06} & 1.07E-02 &       & 1.94E-09 & 3.96E-07 \\
      & SR-Scientist & 7.38E-04 & 6.51E-03 &       & 1.03E-04 & 7.89E-01 &       & 6.54E-06 & \uline{1.14E-03} &       & \textbf{7.68E-14} & \uline{2.20E-10} \\
      & \textbf{Ours} & \textbf{1.95E-06} & \textbf{1.32E-05} &       & \textbf{1.13E-09} & \uline{6.72E-04} &       & \textbf{1.45E-08} & \textbf{6.35E-06} &       & \uline{2.84E-12} & \textbf{2.20E-10} \\
\midrule
Deepseek-v4-flash-0731 & \multirow{5}[2]{*}{\textbf{Ours}} & 1.48E-05 & 8.21E-05 &       & 1.80E-08 & 8.70E-04 &       & 1.75E-07 & 2.25E-05 &       & 3.21E-14 & 1.91E-10 \\
Deepseek-v4.1-flash &       & \textbf{9.71E-07} & \textbf{5.12E-06} &       & \textbf{6.64E-12} & \textbf{7.03E-07} &       & \textbf{1.34E-12} & \textbf{1.32E-08} &       & \uline{2.79E-14} & \textbf{3.51E-12} \\
Deepseek-v4-pro-0813 &       & 6.42E-06 & 1.82E-05 &       & 5.75E-09 & 3.05E-04 &       & 4.84E-09 & 1.38E-06 &       & \textbf{1.92E-14} & \uline{7.50E-12} \\
GLM-5.3-flash &       & \uline{1.95E-06} & \uline{1.32E-05} &       & 1.13E-09 & 6.72E-04 &       & 1.45E-08 & 6.35E-06 &       & 2.84E-12 & 2.20E-10 \\
GLM-5.3 &       & 5.64E-06 & 8.96E-05 &       & \uline{2.48E-10} & \uline{1.13E-05} &       & \uline{4.05E-10} & \uline{1.91E-08} &       & 9.62E-14 & 1.78E-11 \\
\bottomrule
\end{tabular}}
\end{table*}

\begin{table*}[t]
    \centering
    \caption{Runtime, token usage, and API cost on LSR-Synth.}
    \label{tab:lsr-synth-budget}
    \resizeTabular[0.37950008]{\begin{tabular}{llccc}
\toprule
\textbf{Base Model} & \multicolumn{1}{l}{\textbf{Method}} & \textbf{Time (min)} & \textbf{Token (M)} & \textbf{Cost (\$)} \\
\midrule
(no LLM) & \multicolumn{1}{l}{PySR} & 25.46 & /     & / \\
\midrule
\multirow{4}[2]{*}{Deepseek-v4-flash-0731} & \multicolumn{1}{l}{LLM-SR} & 57.67 & \uline{1.00} & 0.05 \\
      & \multicolumn{1}{l}{IGSR} & 24.02 & \textbf{0.125} & \textbf{0.0198} \\
      & \multicolumn{1}{l}{SR-Scientist} & \uline{16.62} & 1.63  & 0.05 \\
      & \multicolumn{1}{l}{\textbf{Ours}} & \textbf{11.11} & 1.53  & \uline{0.03} \\
\midrule
\multirow{4}[2]{*}{GLM-5.3-flash} & \multicolumn{1}{l}{LLM-SR} & 23.06 & \uline{0.28} & 0.09 \\
      & \multicolumn{1}{l}{IGSR} & \textbf{1.43} & \textbf{0.020} & \textbf{0.0023} \\
      & \multicolumn{1}{l}{SR-Scientist} & 16.72 & 0.91  & 0.04 \\
      & \multicolumn{1}{l}{\textbf{Ours}} & \uline{13.07} & 0.912 & \uline{0.0324} \\
\midrule
Deepseek-v4-flash-0731 & \multicolumn{1}{l}{\multirow{5}[2]{*}{\textbf{Ours}}} & \uline{11.11} & 1.53  & \textbf{0.03} \\
Deepseek-v4.1-flash &       & 41.20 & 1.888 & 0.0726 \\
Deepseek-v4-pro-0813 &       & 16.63 & 1.63  & 0.21 \\
GLM-5.3-flash &       & 13.07 & \uline{0.912} & \uline{0.0324} \\
GLM-5.3 &       & \textbf{7.48} & \textbf{0.906} & 0.3582 \\
\bottomrule
\end{tabular}}
\end{table*}

\FloatBarrier
\subsubsection{Backbone Dependence on LSR-Synth}
\label{sec:backbone-analysis}

To examine what properties of the base model predict symbolic-regression performance, we compare the five SRHarness backbones for which all 128 LSR-Synth problems are complete. We exclude the two partial Qwen runs from this analysis. Table~\ref{tab:backbone-analysis} reports aggregate LSR-Synth performance together with model scale, release date, two long-horizon agent benchmarks, and GPQA Diamond as a proxy for scientific question answering~\citep{rein2023gpqa}. Parameter counts and benchmark scores are taken from the corresponding model cards and release documentation~\citep{xu2026deepseek,zeng2026glm}; the GLM-5.3-Flash GPQA value is an independently measured score.

\begin{table*}[h]
    \centering
    \caption{SRHarness performance with five fully evaluated LSR-Synth backbones and selected model attributes. Parameter counts are total/activated parameters per token; DeepSeek-v4.1-flash activates 8B parameters during prefill and 16B during decoding. Benchmark scores use the publicly reported evaluation settings of each model and are therefore descriptive rather than a strictly controlled cross-model evaluation.}
    \label{tab:backbone-analysis}
    \resizeTabular[0.37950008]{    \begin{tabular}{llcccclccc}
        \toprule
        \multirow{2}{*}{\textbf{Method}} &
        \multirow{2}{*}{\textbf{Base Model}} &
        \multicolumn{3}{c}{\textbf{LSR-Synth Performance}} &
        \multirow{2}{*}{\textbf{Parameters}} &
        \multirow{2}{*}{\textbf{Release Date}} &
        \multicolumn{3}{c}{\textbf{Benchmark Score}} \\
        \cmidrule(lr){3-5}\cmidrule(lr){8-10}
        & & \textbf{ID Acc$_{0.1}$ $\uparrow$} & \textbf{ID Median NMSE $\downarrow$} & \textbf{SA $\uparrow$}
        & & & \textbf{TerminalBench 2.1} & \textbf{DeepSWE} & \textbf{GPQA Diamond} \\
        \midrule
        \multirow{5}{*}{Ours}
        & DeepSeek-v4-flash-0731 & 91.51\% & $1.68\times10^{-7}$ & 6.20\% & 284B / 13B & 2026-07-31 & 82.7 & 54.4 & 89.9 \\
        & DeepSeek-v4-pro-0813   & 95.00\% & $5.00\times10^{-9}$ & 11.63\% & 1.6T / 49B & 2026-08-13 & 87.9 & 62.7 & \textbf{92.4} \\
        & DeepSeek-v4.1-flash    & \textbf{95.52\%} & $\mathbf{1.96\times10^{-12}}$ & \textbf{15.50\%} & 552B / 8--16B & 2026-09-10 & \textbf{90.6} & \textbf{74.2} & 90.9 \\
        & GLM-5.3-flash          & 92.56\% & $1.84\times10^{-9}$ & 10.85\% & 320B / 18B & 2026-08-26 & 84.3 & 63.4 & 91.2$^{\dagger}$ \\
        & GLM-5.3                & 93.47\% & $7.58\times10^{-10}$ & 10.85\% & 744B / 40B & 2026-08-14 & 88.2 & 66.9 & 88.1 \\
        \bottomrule
    \end{tabular}    }
    \vspace{1mm}

    \noindent $^{\dagger}$~Independent Artificial Analysis measurement; the other GPQA values are reported in the DeepSeek-v4.1 model card. External benchmark protocols, scaffolds, and reasoning settings are not fully matched.
\end{table*}

\begin{figure*}[t]
    \centering
    \includegraphics[width=\textwidth]{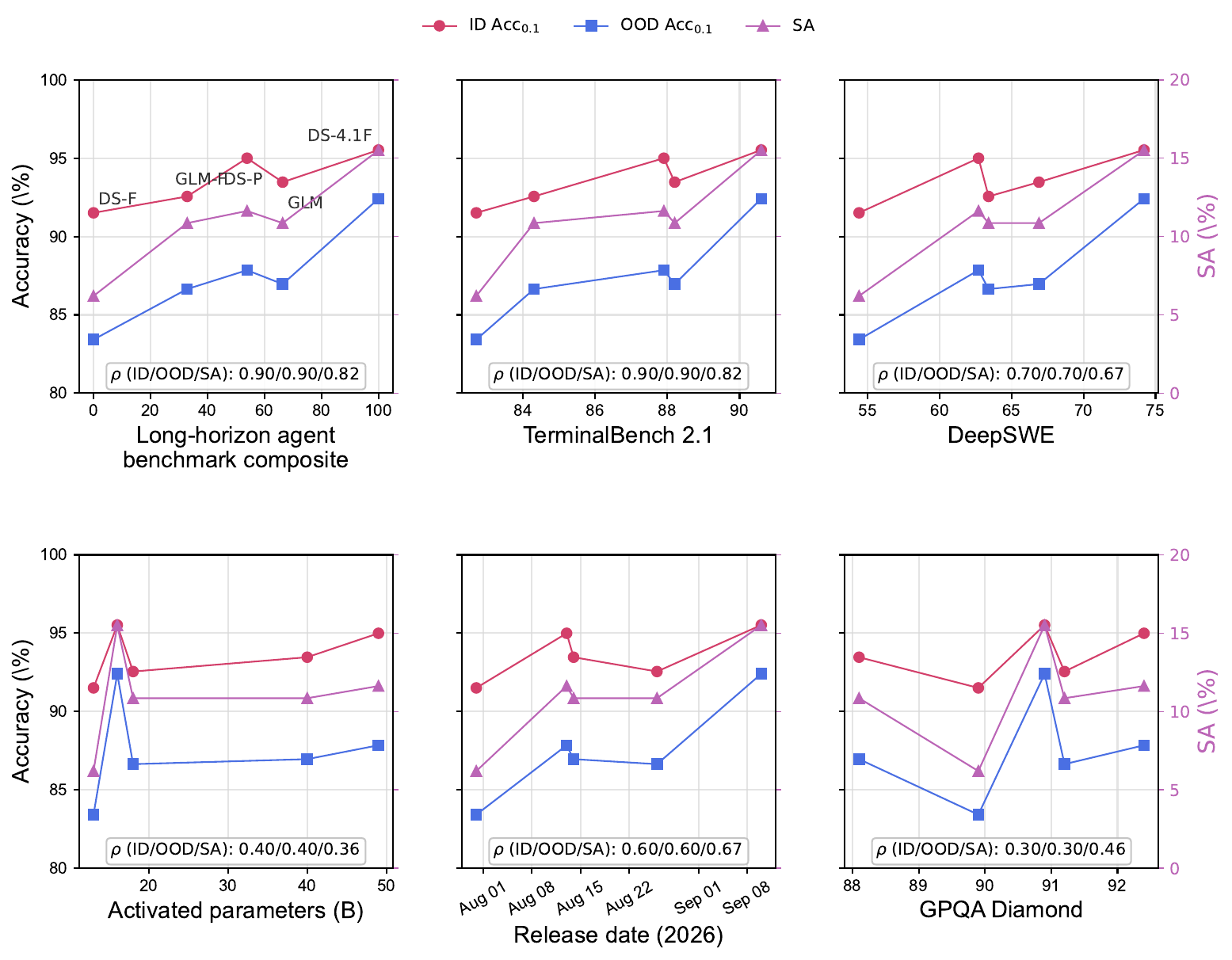}
    \caption{Relationship between backbone attributes and aggregate LSR-Synth performance. Accuracy uses the left axis and SA the right axis, with identical vertical scales in all panels. The first row shows a long-horizon agent composite and its two constituents; the composite averages separately min--max-normalized TerminalBench 2.1 and DeepSWE scores. Activated parameter count is per token and uses the 16B decoding value for DeepSeek-v4.1-flash. Insets report Spearman rank correlations.}
    \label{fig:backbone-performance-correlates}
\end{figure*}

The rankings align much more closely with long-horizon agent benchmarks than with raw model scale (Figure~\ref{fig:backbone-performance-correlates}). Across the five models, TerminalBench 2.1 has Spearman rank correlations of $0.90$, $0.82$, and $0.90$ with overall ID $\mathrm{Acc}_{0.1}$, SA, and $-\log_{10}$ median NMSE, respectively. DeepSWE is especially aligned with numerical fidelity: its ordering---DeepSeek-v4.1-flash, GLM-5.3, GLM-5.3-flash, DeepSeek-v4-pro-0813, and DeepSeek-v4-flash-0731---exactly matches the ordering by median NMSE ($\rho=1.00$). By contrast, total parameter count has correlations of only $0.70$, $0.62$, and $0.30$ with the same three outcomes, while activated parameter count is weaker still ($0.40$, $0.36$, and $0.10$). GPQA also has limited correspondence ($0.30$, $0.46$, and $-0.20$), indicating that static scientific question answering alone does not explain performance in interactive equation discovery.

These comparisons are consistent with the view that performance under SRHarness depends strongly on capabilities required for long-horizon, feedback-driven search, rather than being explained primarily by model scale or static scientific question answering. A successful trajectory requires repeatedly generating hypotheses, selecting analyses and fitting procedures, interpreting numerical feedback, and revising candidate structures over a long interaction. This is consistent with the strong alignment to TerminalBench and DeepSWE. It also explains why scale is not monotonic: DeepSeek-v4.1-flash, with 16B activated parameters during decoding, outperforms the 49B-active DeepSeek-v4-pro-0813 overall, while the two GLM variants obtain identical SA despite their substantial difference in scale.

The aggregate ranking nevertheless hides meaningful domain-specific behavior. DeepSeek-v4.1-flash is substantially stronger on biology, reaching 98.44\% ID and 98.59\% OOD $\mathrm{Acc}_{0.1}$, whereas DeepSeek-v4-pro-0813 reaches 89.51\% and 76.93\%. Conversely, DeepSeek-v4-pro-0813 obtains the highest physics ID/OOD accuracies (93.73\%/92.18\%). GLM-5.3 combines competitive aggregate accuracy with the shortest observed runtime, although wall-clock time also depends on token consumption and provider throughput and should not be interpreted as an intrinsic model-speed measurement. Overall, the evidence suggests that capabilities associated with long-horizon agent behavior and feedback-driven search may be more informative for SRHarness performance than parameter count alone. Because this analysis contains only five models and the external benchmarks use different scaffolds and evaluation protocols, these correlations should be viewed as hypothesis-generating rather than causal evidence.

\FloatBarrier
\subsubsection{Domain-Specific Difficulty in LSR-Synth}
\label{sec:lsrsynth-domain-difficulty}

\begin{figure*}[t]
    \centering
    \includegraphics[width=0.7142857\textwidth]{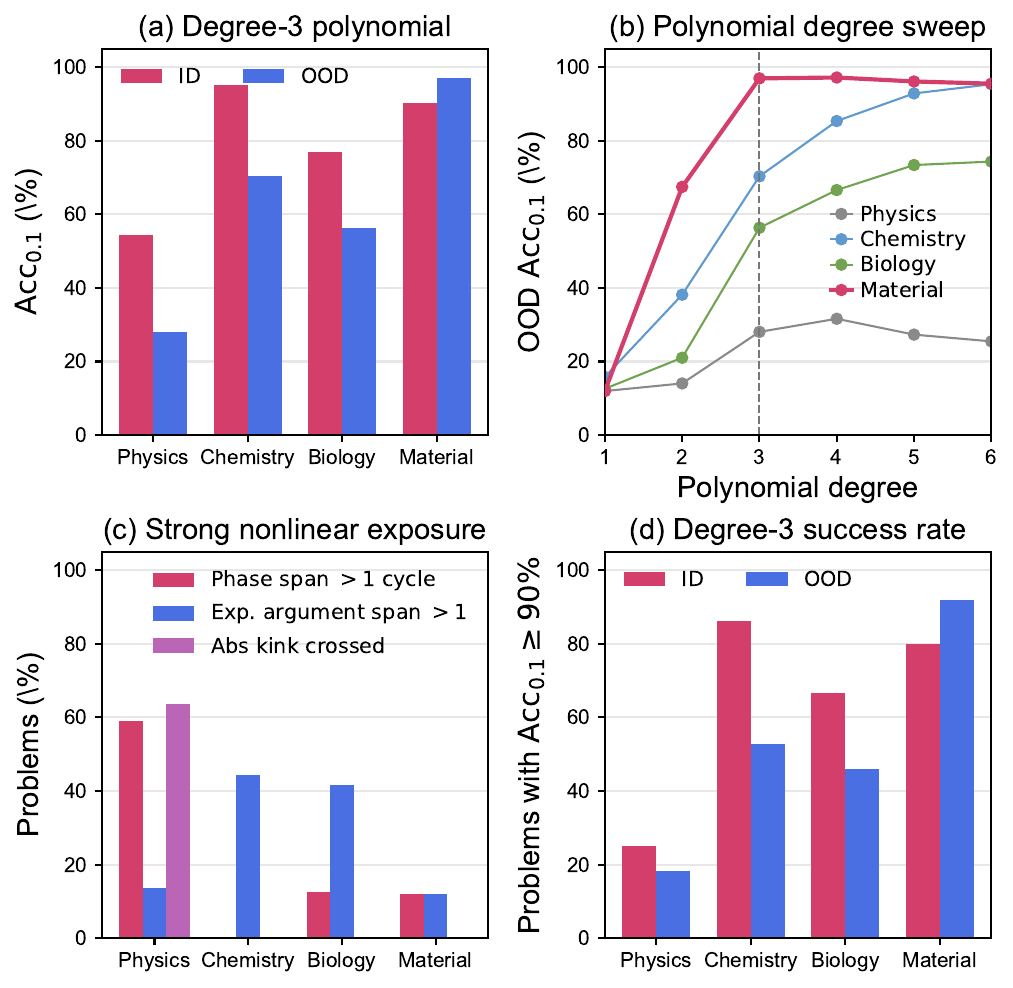}
    \caption{Diagnostics for the differing numerical difficulty of the four LSR-Synth domains. (a) Accuracy of the same degree-three polynomial baseline on every domain. (b) OOD accuracy as polynomial degree increases; the dashed line marks degree three. (c) Fractions of problems with strong nonlinear exposure over the combined train, ID, and OOD range: a periodic argument covering more than one cycle, an exponential argument spanning more than one dimensionless unit, or observations crossing an absolute-value kink. (d) Fractions of individual problems on which the degree-three polynomial exceeds 90\% accuracy.}
    \label{fig:lsrsynth-domain-difficulty}
\end{figure*}

The material-science subset is close to numerical saturation for many methods, and its OOD $\mathrm{Acc}_{0.1}$ can even exceed its ID accuracy. To isolate the source of this behavior, we fit complete polynomials, including all interaction terms, using only the official training observations and evaluate them on the untouched ID and OOD sets with the same numerical pipeline and per-problem aggregation used in the main experiments. Inputs are standardized before fitting for numerical conditioning, which does not change the polynomial function class in the original variables. A degree-three polynomial reaches 90.31\% ID and 96.94\% OOD $\mathrm{Acc}_{0.1}$ on material problems, compared with 54.34\%/28.06\% for physics, 95.02\%/70.32\% for chemistry, and 77.07\%/56.35\% for biology. Moreover, 92\% of material tasks individually exceed 90\% OOD accuracy, versus 18\%, 53\%, and 46\% in the other three domains. The degree sweep in Figure~\ref{fig:lsrsynth-domain-difficulty}(b) shows that material OOD performance already saturates at degree three, whereas chemistry and biology require higher degrees and physics does not extrapolate reliably.

The formula-scale statistics provide a consistent explanation. All material tasks use strain $\epsilon$ and temperature $T$: train and ID cover $\epsilon\in[0,0.54]$ and $T\in[273,543]$ K, while OOD extends them to $\epsilon\in(0.54,0.60]$ and $T\in(543,573]$ K. Four of 25 material expressions are exact polynomials, and most remaining nonlinearities vary weakly over these intervals: the two $\sin(\epsilon)$ terms cover only $0.095$ cycles, $\log(1+\epsilon)$ changes by only $\log(1.6)=0.47$, and the Arrhenius arguments in $\exp[-C/(RT)]$ change by at most 0.022. The median exponential-argument span is consequently 0.016 in material, versus 1.00, 4.24, and 4.95 in physics, chemistry, and biology. Although three temperature-sinusoid and three narrow Gaussian material tasks are harder exceptions, only 12\% of material problems cross more than one periodic cycle and 12\% span more than one exponential-argument unit. By contrast, 26 of 44 physics problems contain $\sin(t)$ over $t\in[0,40]$---6.37 cycles---and 28 of 44 cross an absolute-value kink; chemistry and biology expose 44\% and 42\% of their tasks to an exponential-argument span above one. Thus, across most of the material subset, the observed variable ranges cover a substantially smoother portion of each nonlinear law, making low-degree polynomial approximation unusually effective.

\FloatBarrier
\subsection{Detailed LSR-Transform Results}
\label{sec:lsr-transform-additional-results}

Table~\ref{tab:lsr-transform-budget} provides the runtime, token usage, and API cost associated with the LSR-Transform results in the main text. SRHarness requires only 12.47 minutes per problem under DeepSeek-v4-flash-0731, substantially faster than LLM-SR (72.74 minutes). Notably, this efficiency does not translate into higher cost: despite consuming a comparable number of total tokens (1.143M), SRHarness achieves an API cost of \$0.0204 per problem, far lower than LLM-SR (\$0.0535). This discrepancy is explained by the composition of tokens. SRHarness's calls are dominated by input: prompt tokens constitute approximately 97\% of its total usage, and about 93\% of these prompt tokens hit the provider's prefix cache, while output tokens account for only approximately 3\%. LLM-SR shows the opposite profile: its prompts are short, but every call generates a long program, so most of its usage is billed at the expensive, non-cacheable output rate. Consequently, SRHarness's token-weighted cost (\$0.018/M) is far below LLM-SR's (\$0.053/M) despite its larger total token count. Under GLM-5.3-flash, SRHarness further reduces runtime to 7.09 minutes at \$0.0242 per problem.

\begin{table*}[t]
    \centering
    \caption{Runtime, token usage, and API cost on LSR-Transform.}
    \label{tab:lsr-transform-budget}
    \resizeTabular[0.37950008]{\begin{tabular}{llccc}
\toprule
\textbf{Base Model} & \multicolumn{1}{l}{\textbf{Method}} & \textbf{Time (min)} & \textbf{Token (M)} & \textbf{Cost (\$)} \\
\midrule
(no LLM) & \multicolumn{1}{l}{PySR} & 25.49 & /     & / \\
\midrule
\multirow{4}[2]{*}{Deepseek-v4-flash-0731} & \multicolumn{1}{l}{LLM-SR} & 72.74 & 1.015 & 0.0535 \\
      & \multicolumn{1}{l}{IGSR} & 23.67 & \textbf{0.140} & \uline{0.0224} \\
      & \multicolumn{1}{l}{SR-Scientist} & \textbf{6.72} & \uline{0.9151} & 0.028 \\
      & \multicolumn{1}{l}{\textbf{Ours}} & \uline{12.47} & 1.143 & \textbf{0.0204} \\
\midrule
\multirow{4}[1]{*}{GLM-5.3-flash} & \multicolumn{1}{l}{LLM-SR} & 27.56 & \uline{0.3454} & 0.1176 \\
      & \multicolumn{1}{l}{IGSR} & \textbf{1.58} & \textbf{0.024} & \textbf{0.0027} \\
      & \multicolumn{1}{l}{SR-Scientist} & 7.41  & 0.4032 & \uline{0.0204} \\
      & \multicolumn{1}{l}{\textbf{Ours}} & \uline{7.09} & 0.690 & 0.0242 \\
\midrule
Deepseek-v4-flash-0731 & \multicolumn{1}{l}{\multirow{4}[1]{*}{\textbf{Ours}}} & 12.47 & 1.143 & \textbf{0.0204} \\
Deepseek-v4-pro &       & 13.77 & 1.272 & 0.2486 \\
GLM-5.3-flash &       & \uline{7.09} & \uline{0.690} & \uline{0.0242} \\
GLM-5.3 &       & \textbf{3.57} & \textbf{0.666} & 0.2325 \\
\bottomrule
\end{tabular}}
\end{table*}

\FloatBarrier
\subsection{Detailed LSR-Transform-Anon Results}
\label{sec:lsr-transform-anon-additional-results}

Table~\ref{tab:lsr-transform-anon-budget} reports the corresponding resource usage on LSR-Transform-Anon. SRHarness requires only 12.34 minutes and \$0.0252 per problem under DeepSeek-v4-flash-0731, substantially outperforming LLM-SR (286.13 minutes, \$0.1187) and SR-Scientist (22.30 minutes, \$0.0449). Notably, this cost and runtime remain essentially unchanged from the standard LSR-Transform setting (12.47 minutes, \$0.0204), indicating that removing semantic priors does not increase the search burden for SRHarness. By contrast, several baselines require drastically longer trajectories to recover far fewer target equations, corroborating the main-text finding that structured tool-mediated exploration partially compensates for the loss of semantic cues. This efficiency is robust across backbones, with GLM-5.3-flash yielding 15.15 minutes and \$0.0310 per problem.

\begin{table*}[t]
    \centering
    \caption{Runtime, token usage, and API cost on LSR-Transform-Anon.}
    \label{tab:lsr-transform-anon-budget}
    \resizeTabular[0.37950008]{\begin{tabular}{llccc}
\toprule
\textbf{Base Model} & \multicolumn{1}{l}{\textbf{Method}} & \textbf{Time (min)↓} & \textbf{Token (M)} & \textbf{Cost (\$)} \\
\midrule
(no LLM) & \multicolumn{1}{l}{PySR} & 25.49 & /     & / \\
\midrule
\multirow{4}[2]{*}{Deepseek-v4-flash-0731} & \multicolumn{1}{l}{LLM-SR} & 286.13 & 2.253 & 0.1187 \\
      & \multicolumn{1}{l}{IGSR} & 55.60 & \textbf{0.163} & \uline{0.0269} \\
      & \multicolumn{1}{l}{SR-Scientist} & \uline{22.30} & 1.587 & 0.0449 \\
      & \multicolumn{1}{l}{\textbf{Ours}} & \textbf{12.34} & \uline{1.300} & \textbf{0.0252} \\
\midrule
Deepseek-v4-flash-0731 & \multicolumn{1}{l}{\multirow{3}[2]{*}{\textbf{Ours}}} & \textbf{12.34} & \uline{1.300} & \textbf{0.0252} \\
Deepseek-v4-pro &       & 17.30 & 1.508 & 0.2862 \\
GLM-5.3-flash &       & \uline{15.15} & \textbf{0.883} & \uline{0.0310} \\
\bottomrule
\end{tabular}}
\end{table*}

\FloatBarrier
\subsubsection{Codex Evaluation Protocol and Additional Comparison}
\label{sec:codex-anon-additional-results}

We evaluate Codex as a single-agent solver by launching a fresh Codex CLI process for each problem in an isolated, workspace-write sandbox with a 900-second wall-clock limit and no sub-agents. Its public workspace contains the official training observations, agent-facing variable metadata, and instructions to write one discovered expression to a designated result file. It does not contain the reference expression or the benchmark test observations. Standard LSR-Transform supplies the original variable names and descriptions, whereas LSR-Transform-Anon supplies only $y$, $x_1,x_2,\ldots$ and generic input/output labels. The DeepSeek-v4-flash-0731 comparison uses Codex CLI v0.152.0 through an OpenRouter profile; the GPT-5.5 runs invoked an unpinned \texttt{@openai/codex@latest}, whose exact historical CLI version was not retained. In the no-tools condition, the SRHarness scientific-action wrapper is disabled, but Codex retains its native sandboxed terminal and can write Python analyses over the training data. In the tools condition, the same Codex controller additionally receives the SRHarness scientific actions; interactive human assistance and unrestricted workspace-shell actions remain disabled.

Table~\ref{tab:codex-budget} gives the resource comparison between SRHarness and the Codex configurations evaluated on LSR-Transform-Anon. Under the matched DeepSeek-v4-flash-0731 backbone, SRHarness and Codex consume comparable time (12.34 vs. 12.81 min) and token counts (1.300M vs. 1.146M), yet SRHarness achieves an SA of 72.97\% versus 20.72\% for Codex (main text Table 4). Furthermore, directly exposing SRHarness's scientific tools to Codex with GPT-5.5 increases both runtime (5.54 to 7.47 min) and token consumption (0.618M to 1.231M), while its SA actually drops from 68.47\% to 64.86\%. These statistics corroborate the main-text argument that a harness is not merely a collection of tools, and that enlarging the action space without proper integration can introduce unnecessary search burden.

\begin{table*}[t]
    \centering
    \caption{Resource usage for the SRHarness--Codex comparison on LSR-Transform-Anon.}
    \label{tab:codex-budget}
    \resizeTabular[0.37950008]{\begin{tabular}{rlccc}
\toprule
\multicolumn{1}{r}{\multirow{2}[4]{*}{\textbf{Base Model}}} & \multirow{2}[4]{*}{\textbf{Harness}} & \multicolumn{3}{c}{\textbf{LSR-Transform (Anonymized)}} \\
\cmidrule{3-5}      &       & \textbf{Time (min)} & \textbf{Token (M)} & \textbf{Cost (\$)} \\
\midrule
\multicolumn{1}{r}{\multirow{2}[2]{*}{Deepseek-v4-flash-0731}} & \textbf{Ours} & \textbf{12.34} & 1.300 & \textbf{0.0252} \\
      & Codex & 12.81 & \textbf{1.146} & 0.0333 \\
\midrule
\multicolumn{1}{r}{\multirow{2}[2]{*}{GPT-5.5}} & Codex & \textbf{5.54} & \textbf{0.618} & / \\
      & Codex (+tools) & 7.47  & 1.231 & / \\
\bottomrule
\end{tabular}}
\end{table*}

\FloatBarrier
\subsection{Detailed Ablation Results}
\label{sec:ablation-details}

We validate the three design dimensions of SRHarness on LSR-Transform-Anon with DeepSeek-v4-flash-0731. We want to show that the agent must analyze numerical observations, maintain competing hypotheses, and refine them using intermediate evidence. The following results complement Section~\ref{sec:ablation} by relating symbolic recovery to numerical fidelity, expression complexity, and resource usage.

\subsubsection{Repeated Runs}

Table~\ref{tab:ablation-repeat} lists the three runs and their average used to define the full-configuration ablation reference. 
SA ranges from 72.97\% to 75.68\%, with an average of 74.47\%, while numerical accuracy remains between 91.69\% and 92.77\%. Expression complexity varies from 17.55 to 33.36, despite all three runs achieving median NMSE on the order of $10^{-14}$. Thus, similar numerical fidelity can accompany different discovered structures, consistent with the distinction between fitting observations and recovering symbolic laws in the main experiments. 

\begin{table*}[!htbp]
    \centering
    \caption{Variation across the three runs used for the full-configuration ablation reference.}
    \label{tab:ablation-repeat}
    \resizeTabular[0.37950008]{\begin{tabular}{rlccccccc}
\toprule
\multicolumn{1}{r}{\multirow{2}[4]{*}{\textbf{Base Model}}} & \multirow{2}[4]{*}{\textbf{Harness}} & \multicolumn{7}{c}{\textbf{LSR-Transform (Anonymized)}} \\
\cmidrule{3-9}      &       & \textbf{Test Acc0.1 (\%)↑} & \textbf{Test NMSE↓} & \textbf{SA (\%) ↑} & \textbf{Complexity↓} & \textbf{Time (min)} & \textbf{Token (M)} & \textbf{Cost (\$)} \\
\midrule
\multicolumn{1}{r}{\multirow{4}[4]{*}{Deepseek-v4-flash-0731}} & \textbf{Ours (run 1)} & 91.69\% & 1.24E-14 & 72.97\% & 33.36 & \textbf{12.34} & \textbf{1.300} & \textbf{0.0252} \\
      & \textbf{Ours (run 2)} & \uline{92.30\%} & \textbf{1.09E-14} & \uline{74.77\%} & \uline{18.67} & 15.92 & 1.379 & 0.0403 \\
      & \textbf{Ours (run 3)} & \textbf{92.77\%} & \textbf{1.09E-14} & \textbf{75.68\%} & \textbf{17.55} & 16.55 & 1.382 & 0.0404 \\
\cmidrule{2-9}      & \textbf{Ours (3 runs average)} & 92.25\% & \uline{1.24E-14} & 74.47\% & 23.19 & \uline{14.93} & \uline{1.354} & \uline{0.0353} \\
\bottomrule
\end{tabular}}
\end{table*}

\FloatBarrier
\subsubsection{Ablation of Scientific Actions}

Table~\ref{tab:ablation-action} reports the complete numerical, symbolic, complexity, and resource metrics for the scientific-action variants.
Restricting SRHarness to the code executor reduces SA from 74.47\% to 68.47\%, although mean search time changes only from 14.93 to 14.34 minutes. This supports the value of structured scientific actions beyond general computation. Such support is especially relevant when scientific descriptions are unavailable and search must be guided by evidence extracted from the data.

Removing expression-based views reduces SA to 67.57\%, even though median NMSE remains near $10^{-14}$. These views allow heterogeneous operations to act on transformed variables and candidate-derived quantities so that a current hypothesis can guide subsequent analysis. The decline is consistent with our motivation for composable scientific actions: effective discovery requires not only applying tools to the original observations, but also reusing intermediate evidence to refine an evolving hypothesis. 

Adding fitting actions increases SA to 79.28\%, illustrating how the shared action interface can accommodate capabilities that benefit discovery. However, the mean complexity increases from 23.19 to 33.16, and both runtime and token usage increase. These results support extensible scientific actions while showing that their value must be assessed through the resulting search behavior, rather than only the tool availability.

\begin{table*}[!htbp]
    \centering
    \caption{Ablation of the scientific action space on LSR-Transform-Anon.}
    \label{tab:ablation-action}
    \resizeTabular{\begin{tabular}{rlccccccc}
\toprule
\multicolumn{1}{r}{\multirow{2}[3]{*}{\textbf{Base Model}}} & \multirow{2}[3]{*}{\textbf{Harness}} & \multicolumn{7}{c}{\textbf{LSR-Transform (Anonymized)}} \\
\cmidrule{3-9}      &       & \textbf{Test Acc0.1 (\%)↑} & \textbf{Test NMSE↓} & \textbf{SA (\%) ↑} & \textbf{Complexity↓} & \textbf{Time (min)} & \textbf{Token (M)} & \textbf{Cost (\$)} \\
\midrule
\multicolumn{1}{r}{\multirow{4}[1]{*}{Deepseek-v4-flash-0731}} & \textbf{Ours (Full, 3 runs average)} & \uline{92.25\%} & \uline{1.24E-14} & \uline{74.47\%} & \textbf{23.19} & 14.93 & 1.354 & 0.0353 \\
      & \textbf{Ours (Code executor only)} & 88.15\% & 1.66E-14 & 68.47\% & 26.39 & \uline{14.34} & \textbf{1.124} & \uline{0.0349} \\
      & \textbf{Ours (Additional fitting actions)} & \textbf{92.39\%} & \textbf{1.07E-14} & \textbf{79.28\%} & 33.16 & 17.39 & 1.449 & 0.0419 \\
      & \textbf{Ours (w/o Expression-based views)} & 90.29\% & 1.54E-14 & 67.57\% & \uline{25.07} & \textbf{11.59} & \uline{1.238} & \textbf{0.0311} \\
\bottomrule
\end{tabular}}
\end{table*}

\FloatBarrier
\subsubsection{Ablation of Persistent State}

Table~\ref{tab:ablation-state} reports the complete results for the alternative model-facing views of persistent scientific state.
Presenting only the best formula reduces SA from 74.47\% to 66.67\%, with little reduction in token usage and a longer mean runtime. By contrast, removing the explicit best-formula view retains 72.97\% SA. 

The variant without the explicit best-formula view also achieves higher numerical accuracy and lower mean complexity than the full configuration, while its SA is slightly lower. These metrics capture different aspects of discovery, so neither numerical fit nor compactness alone establishes better symbolic recovery. The results show that trajectory lifecycle management needs a trade-off among refinement, exploration, and execution cost.

\begin{table*}[!htbp]
    \centering
    \caption{Ablation of the model-facing scientific-state view on LSR-Transform-Anon.}
    \label{tab:ablation-state}
    \resizeTabular{\begin{tabular}{rlccccccc}
\toprule
\multicolumn{1}{r}{\multirow{2}[3]{*}{\textbf{Base Model}}} & \multirow{2}[3]{*}{\textbf{Harness}} & \multicolumn{7}{c}{\textbf{LSR-Transform (Anonymized)}} \\
\cmidrule{3-9}      &       & \textbf{Test Acc0.1 (\%)↑} & \textbf{Test NMSE↓} & \textbf{SA (\%) ↑} & \textbf{Complexity↓} & \textbf{Time (min)} & \textbf{Token (M)} & \textbf{Cost (\$)} \\
\midrule
\multicolumn{1}{r}{\multirow{3}[1]{*}{Deepseek-v4-flash-0731}} & \textbf{Ours (Pareto view, 3 runs average)} & \uline{92.25\%} & \textbf{1.24E-14} & \textbf{74.47\%} & \uline{23.19} & \textbf{14.93} & 1.354 & \textbf{0.0353} \\
      & \textbf{Ours (Best-formula view)} & 90.84\% & 1.61E-14 & 66.67\% & 25.75 & 17.49 & \uline{1.341} & 0.0400 \\
      & \textbf{Ours (w/o Best-formula view)} & \textbf{93.12\%} & \uline{1.24E-14} & \uline{72.97\%} & \textbf{19.56} & \uline{15.04} & \textbf{1.297} & \uline{0.0385} \\
\bottomrule
\end{tabular}}
\end{table*}

\FloatBarrier
\subsubsection{Ablation of Trajectory Management}

Table~\ref{tab:ablation-lifecycle} reports the complete results for the alternative trajectory-budget allocations.
Relative to the default R1-C1-L30-K1 configuration, the alternatives halve the refinement depth and allocate the remaining budget to another trajectory dimension. They test scheduling choices supported by the lifecycle layer, rather than the presence or absence of lifecycle management.

More restarts and more independent branches yield 66.67\% and 72.07\% SA, respectively, compared with the default 74.47\%. We think that shorter trajectories leave fewer steps to refine hypotheses using accumulated evidence. Although persistent scientific state preserves candidates across restarts, retrieving those candidates does not reproduce the full conversational context in which they were developed. Independent branches may repeat earlier exploration instead of extending a promising line of refinement. Under the tested budget, allocating more capacity to fresh trajectories does not compensate for reduced refinement depth.

Local response sampling achieves 73.87\% SA, within the range of the three default runs, while improving numerical accuracy to 95.38\% and reducing mean complexity to 16.30. Comparing alternative continuations under the same accumulated scientific context may help select more accurate and compact candidates, but these results do not establish improved symbolic recovery. The results show that trajectory lifecycle management needs a trade-off among refinement, exploration, and execution cost.

\begin{table*}[!htbp]
    \centering
    \caption{Ablation of trajectory-budget allocation on LSR-Transform-Anon.}
    \label{tab:ablation-lifecycle}
    \resizeTabular{\begin{tabular}{rlccccccc}
\toprule
\multicolumn{1}{r}{\multirow{2}[3]{*}{\textbf{Base Model}}} & \multirow{2}[3]{*}{\textbf{Harness}} & \multicolumn{7}{c}{\textbf{LSR-Transform (Anonymized)}} \\
\cmidrule{3-9}      &       & \textbf{Test Acc0.1 (\%)↑} & \textbf{Test NMSE↓} & \textbf{SA (\%) ↑} & \textbf{Complexity↓} & \textbf{Time (min)} & \textbf{Token (M)} & \textbf{Cost (\$)} \\
\midrule
\multicolumn{1}{r}{\multirow{4}[1]{*}{Deepseek-v4-flash-0731}} & \textbf{Ours (Default, R1-C1-L30-K1, 3 runs average)} & \uline{92.25\%} & 1.24E-14 & \textbf{74.47\%} & 23.19 & \textbf{14.93} & 1.354 & \textbf{0.0353} \\
      & \textbf{Ours (More restarts, R2-C1-L15-K1)} & 91.34\% & \uline{1.14E-14} & 66.67\% & \uline{20.35} & 27.60 & \uline{1.050} & 0.0379 \\
      & \textbf{Ours (More independent branches, R1-C2-L15-K1)} & 91.80\% & 1.24E-14 & 72.07\% & 22.09 & \uline{22.98} & \textbf{0.950} & \uline{0.0358} \\
      & \textbf{Ours (Local response sampling, R1-C1-L15-K2)} & \textbf{95.38\%} & \textbf{9.76E-15} & \uline{73.87\%} & \textbf{16.30} & 23.73 & 1.296 & 0.0411 \\
\bottomrule
\end{tabular}}
\end{table*}

\FloatBarrier

\end{document}